\documentclass[a4paper,fleqn]{cas-dc}

\usepackage[numbers,sort]{natbib}
\usepackage{amsfonts}
\usepackage {amsmath}
\usepackage{algorithm}
\usepackage{algorithmic}
\usepackage{graphicx}
\usepackage{epstopdf}
\usepackage{caption,subcaption}
\usepackage{booktabs}
\usepackage{array}
\usepackage{newtxtext}
\usepackage{color}
\usepackage{multirow}
\def\tsc#1{\csdef{#1}{\textsc{\lowercase{#1}}\xspace}}
\tsc{WGM}
\tsc{QE}
\tsc{EP}
\tsc{PMS}
\tsc{BEC}
\tsc{DE}
\graphicspath{{pix/}}

\newcommand{\dr}[1]{{\color{black}{#1}}}  
\newcommand{\dt}[1]{\iffalse{#1}\fi}  

\begin{document}

\captionsetup[figure]{name={Fig.},labelsep=period}

\let\WriteBookmarks\relax
\def\floatpagepagefraction{1}
\def\textpagefraction{.001}
\shorttitle{Exploring the Intrinsic Probability Distribution for Hyperspectral Anomaly Detection}
\shortauthors{Yu et~al.}

\title [mode = title]{Exploring the Intrinsic Probability Distribution for Hyperspectral Anomaly Detection}  

\author[1]{Shaoqi Yu}
\author[1]{Xiaorun Li}
\fnmark[*]
\author[1]{Shuhan Chen}
\author[2]{Liaoying Zhao}

\address[1]{Zhejiang University, Colledge of Electrical Engineering, 38 Zheda Rd, Hangzhou, China, 310027}
\address[2]{Hangzhou Dianzi University, Colledge of Computer Science, Xiasha Higher Education Zone, Hangzhou, China, 310018}

\cortext[cor1]{Corresponding author(email:lxr@zju.edu.cn)}

\begin{abstract}
%
%
In recent years, neural network-based anomaly detection methods have attracted considerable attention in the hyperspectral remote sensing domain due to the powerful reconstruction ability compared with traditional methods. However, actual probability distribution statistics hidden in the latent space are not discovered by exploiting the reconstruction error because the probability distribution of anomalies is not explicitly modeled. To address the issue, we propose a novel probability distribution representation detector (PDRD) that explores the intrinsic distribution of both the background and the anomalies in original data for hyperspectral anomaly detection in this paper. First, we represent the hyperspectral data with multivariate Gaussian distributions from a probabilistic perspective. Then, we combine the local statistics with the obtained distributions to leverage the spatial information. Finally, the difference between the corresponding distributions of the test pixel and the average expectation of the pixels in the Chebyshev neighborhood is measured by computing the modified Wasserstein distance to acquire the detection map. We conduct the experiments on four real data sets to evaluate the performance of our proposed method. Experimental results demonstrate the accuracy and efficiency of our proposed method compared to the state-of-the-art detection methods.
\end{abstract}




\maketitle

\section{Introduction}

%
%
%

Hyperspectral imagery (HSI) captures hundreds of continuous narrow bands covering a wide range of wavelengths to record the spectral signature with abundant information \cite{8833518}. The intensity of a specific object in the hyperspectral image reflects the reflectance or radiance that mainly depends on the category to which the object belongs \cite{974718}. Therefore, it is possible to apply the unique spectral characteristic to distinguish numerous ground objects \cite{8945176}. With the improvement of hyperspectral technological sensors of spectral resolution \cite{6555921}, HSI has been involved in multiple domains such as aerial reconnaissance \cite{1220247}, military surveillance \cite{6723732}, and marine monitoring \cite{9104925}, where hyperspectral anomaly detection acts as a critical technology \cite{tu2020hyperspectral}.

Hyperspectral anomaly detection is considered a particular case of target detection \cite{7476876}, \cite{9082140}. It requires no prior knowledge for background or specific objects compared to target detection tasks \cite{974730}, \cite{6678247}. As a consequence, it owns a promising application prospect. The leading hyperspectral anomaly detection theory considers that the objects in HSI can be categorized into the background component and the anomaly component \cite{7332782}, \cite{1020263}, \cite{8421593}. Thus, it is rational to model hyperspectral anomaly detection as a binary classification problem. The object whose spectral signature is significantly dissimilar to their adjacent local neighborhood is regarded as an anomaly \cite{5546306}. The objects in most of the continuous areas belong to the background, while anomalies only occupy a small portion of the image \cite{6810158}. Hence, the distribution of anomalies is sparse \cite{7043320}. 

Traditional hyperspectral anomaly detection methods pay much attention to the characteristic of background \cite{7119558}, \cite{6327353}. The landmark Reed-Xiaoli (RX) detector \cite{60107} holds the assumption that the whole background obeys a  multivariate Gaussian distribution. Based on the generalized likelihood ratio test, the Mahalanobis distance between the test pixel and the mean value of background distribution is calculated. When the distance is greater than a specific threshold, the hypothesis that the test pixel belongs to an anomaly can be verified. Two typical versions of  RX, namely global RX (GRX) detector and local RX (LRX) detector \cite{6412738}, evaluate the background by global statistics and local statistics in HSI, respectively. However, it is difficult for all types of background objects to conform to a single normal distribution from a more general viewpoint. To overcome the limit, Kernel RX (KRX) \cite{1386510} was proposed to model the original data with a nonGaussian model in high dimensional feature space. However, the computational cost remains a troublesome issue to be optimized. 

Recently, with the development of compressed sensing technology in the signal processing domain \cite{1614066}, hyperspectral anomaly detection methods have been developed based on representation learning \cite{6472238} and matrix decomposition theories \cite{doi:10.1137/080738970}. Collaborative representation detector (CRD) \cite{6876207} utilizes the conclusion that the background objects can be approximately represented by their local neighborhood, while the anomalies can not. Thus the reconstruction error is leveraged to distinguish the anomalies from the background objects. Moreover, Robust PCA (RPCA) \cite{10.1117/1.JRS.8.083641} method separates the original hyperspectral data into a background component and an anomaly component by implementing a simple matrix decomposition procedure \cite{NIPS2009_3704}, therein the anomaly component is used for the following detection task. The method assumes that the background objects lie in a single subspace with no complicated implicit structure, whereas it is not realistic in most hyperspectral scenes. To surmount the constraint, low rank and sparse representation (LRASR) model \cite{7322257} was proposed by incorporating the low rank representation (LRR) theory \cite{6180173}  into the sparse representation detector (SRD) \cite{5704546}. LRASR employs the concept of background dictionary, which considers potential extensive categories of background objects. The nuclear norm constraint is added to the coefficients matrix of background to acquire the lowest rank representation of entire spectra jointly, and $l_{\text{2,1}}$ constraint is imposed on the anomaly part to make the distribution of anomalies sparse. However, the detection performance of the LRASR method depends largely on the selection of the background dictionary. Furthermore, low rank and sparse matrix decomposition (LSMAD) detector  \cite{7293169} inserts a noise term in the matrix decomposition procedure to alleviate the contamination simultaneously. Then, the Mahalanobis distance is computed by using the background statistics to obtain the final detection result. In addition, Li \textit{et al.} \cite{9011733} integrates the variational Bayes with the low-rank and sparse decomposition model. The anomaly component can be easily separated from the noise, which is modeled by a mixture of Gaussian.

Apart from the aforementioned methods, novel detectors have been proposed based on neural networks in recent years. The convolutional neural network (CNN) based model \cite{7875485} trains CNN by constructing the training sample pairs. Moreover, Autoencoder (AE) based framework has become prevalent due to the simplicity. Beti \textit{et al.} \cite{10.1117/12.2195180} proposed a semi-supervised learning hyperspectral anomaly detection algorithm based on AE. The model trains the neural network with background samples according to the known ground truth. Thus the anomaly pixels produce large reconstruction errors to separate them from the background objects. However, regardless of AE-based or CNN based models, no consideration is attached to local spatial information that makes a big difference in hyperspectral anomaly detection. To address this issue, Lu \textit{et al.} \cite{8889706} incorporated the embedding manifold constraint into AE to sufficiently utilize the local structural information. Besides,  \cite{an2015variational} is the first method to apply variational autoencoder (VAE) \cite{2013arXiv1312.6114K} to anomaly detection tasks, in which variational inference is integrated with AE in the form of probability distribution. Despite the high reconstruction ability of VAE \cite{9057548}, the intrinsic probability distribution cannot be thoroughly explored. 

Most hyperspectral anomaly detection methods rarely consider the explicit distribution of the anomalies. RX-based approaches model the background as a multivariate Gaussian distribution, but the distribution of anomalies is ignored. Ordinary AE and VAE based methods exploit the reconstruction error to distinguish the anomalies from the background. However, intrinsic distribution information is not explored because neither VAE nor AE models the distribution of anomalies. Moreover, some of them work in a semi-supervised manner, which indicates the ground truth needs to be known in advance. Hence, the motivation of this paper is to discover the probabilistic properties of the hyperspectral data from a probabilistic perspective and explore a valid representation that works in an unsupervised manner with higher performance of generalization. 

In this paper, we propose a novel probability distribution representation detector (PDRD) based on VAE structure for hyperspectral anomaly detection without using reconstruction error, which explicitly represents the HSI with multivariate Gaussian distributions and detects the anomalies by employing the modified Wasserstein distance. The VAE structure is adopted to acquire the representation of training samples in a probabilistic modality. The outputs of the encoder, including the mean value vector and the standard deviation vector, jointly constitute a multivariate Gaussian distribution in the latent feature space. Then we integrate the spatial information with the obtained distribution for each pixel by introducing the concept of Chebyshev neighborhood. For each pixel in the spatial space with its corresponding probability distribution, we define its Chebyshev neighborhood and compute the average expectation distribution of the pixels in the neighborhood to estimate the local statistics. Finally, we employ the modified Wasserstein distance to evaluate the difference between the corresponding distribution of the test pixel and the average expectation distribution that corresponds to the pixels in the neighborhood. In this way, we combine both spectral and spatial information to obtain the final detection results.

The main contributions of the paper can be concluded as follows.

1) We propose a framework to represent both the background and the anomalies in HSI by multivariate Gaussian distributions, which can discover the probabilistic characteristic of the original data in the latent space.

2) Instead of exploiting reconstruction error, we integrate local statistics with probabilistic structural information by constructing the Chebyshev neighborhood for each pixel.

3) We build a valid criterion according to the actual property of HSI to evaluate the difference between two probability distributions, which highlights the anomalies and suppresses the background pixels.

The remainder of this paper is organized as follows. Section \uppercase\expandafter{\romannumeral2}  briefly introduces the related work. In Section \uppercase\expandafter{\romannumeral3}, we elaborate on our proposed method. Experimental results on four data sets are presented in Section \uppercase\expandafter{\romannumeral4}. Finally, Section \uppercase\expandafter{\romannumeral5} draws the conclusion.
\section{Related Work}


The VAE model is a typical generative neural network, which has been broadly applied in computer vision tasks recently. It estimates the probability distribution of the training data and then samples examples from the learned probability distribution, aiming to generate new images that look similar to the original data. Different from the original AE framework, VAE views the reconstruction problem from a probabilistic perspective. 

For each sample  $ \mathbf{x} $, the goal of VAE is to maximize $ P\mathbf{(x)}$. According to the law of total probability, $  P\mathbf{(x)}$ is represented as

\begin{equation}
	P \mathbf{(x)}=\int_{z} P(\mathbf{x|z}) P \mathbf{(z)} dz
\end{equation}
where $  \mathbf{z}$ indicates the latent variable to describe the implicit probability distribution. $P(\mathbf{x|z})$ is intractable in the actual calculation procedure, thus variational inference is imposed to solve the dilemma. The loss function of  VAE is then modeled by
\begin{equation}
	\mathcal{L}=\frac{1}{N} \sum_{i=1}^{N} (E_{q_{\phi}(\mathbf{z} | \mathbf{x}_{i})}\left[\log p_{\theta}(\mathbf{x}_{i} | \mathbf{z})\right]-\mathbb{KL}(q_{\phi}(\mathbf{z} | \mathbf{x}_{i}) \| p_{\theta}(\mathbf{z})))
	\label{loss function}
\end{equation}
where $\phi$ and $\theta$ denotes the parameters of the encoder and decoder, respectively. $N$ indicates the total number of training samples. The first term is called reconstruction likelihood, which measures the log-likelihood $ \log p_{\theta}(\mathbf{x}_{i}| \mathbf{z})$  of sampling from $q_{\phi}(\mathbf{z} | \mathbf{x}_{i})$. The second KL-divergence term is usually called complexity constraint. It avoids the estimated posterior $q_{\phi}(\mathbf{z} | \mathbf{x}_{i})$ deviating from the prior $p_{\theta}(\mathbf{z})$ too much. The objective function of VAE ensures reconstruction accuracy, along with the generative ability of the model.

As an anomaly detection method using reconstruction error, VAE is more practical than AE due to prominent representational ability. For a given sample $\mathbf{x}$, the probabilistic encoder generates the mean value vector and standard deviation vector in the latent feature space. To reconstruct  $\mathbf{x}$ from the latent space, these two vectors are combined with a random term and then fed into the decoder network. The reconstruction error is employed for the following anomaly detection process. For traditional images, VAE has achieved a satisfactory effect on anomaly detection problems  \cite{2018arXiv180701349L}, \cite{kiran2018overview}.

Ordinary neural network-based methods such as AE and VAE  weigh the reconstruction error as a vital term in anomaly detection tasks. However, the difference between the anomalies and the background can be discriminated through the reconstruction error when training samples contain only background pixels.  In other words, these methods that depend on reconstruction error require the ground truth of the image to be known in advance. Neither AE nor VAE can serve as an unsupervised method, which heavily limits the practical application.
\section{Methodology}
\begin{figure*}
	\centering
	\includegraphics[width=\linewidth]{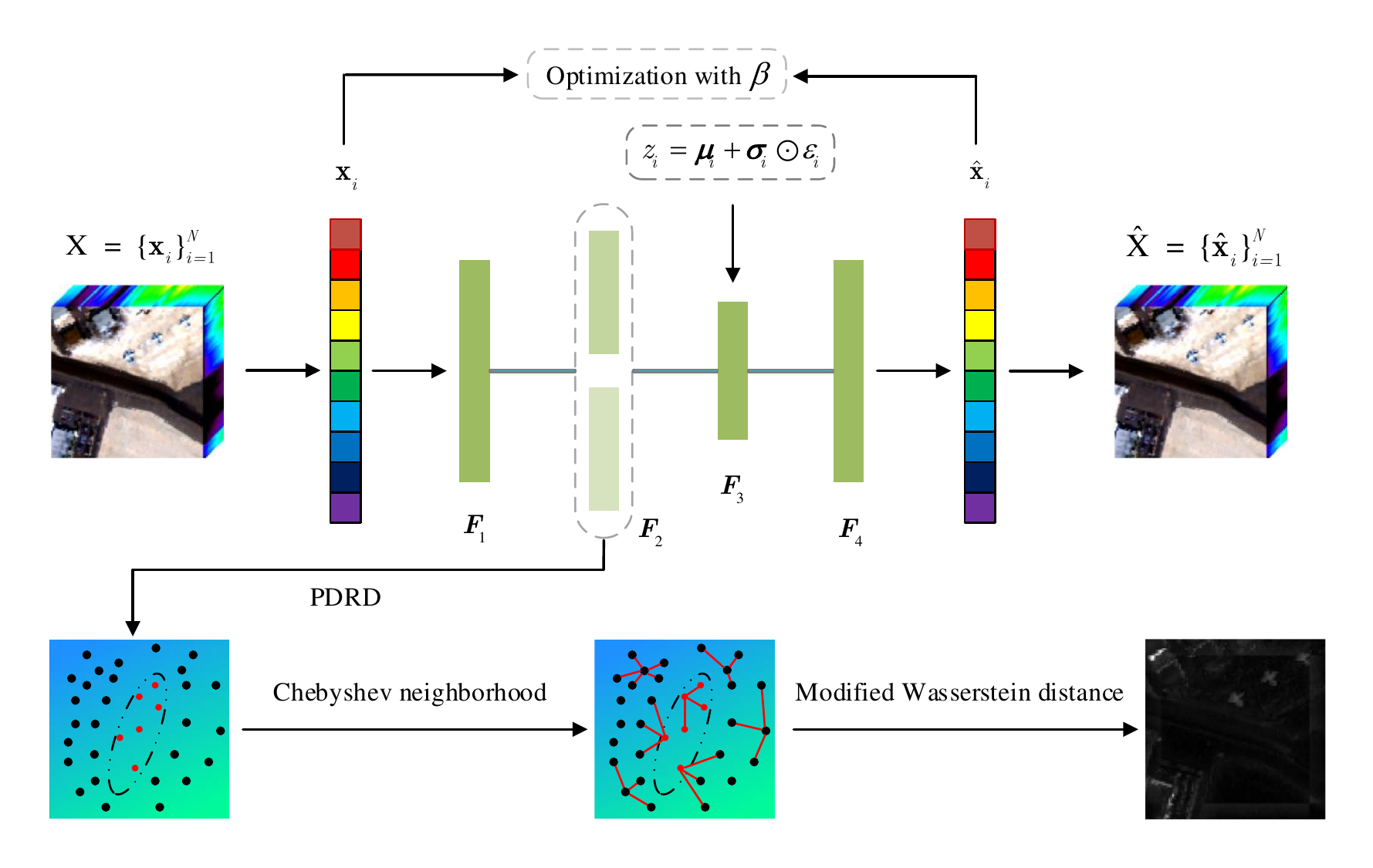}\\
	\caption{Framework of the proposed method.}
	\label{Framework}
\end{figure*}

In this section, we elaborate on our proposed method based on the VAE architecture, which is implemented in an unsupervised learning manner. As illustrated in Fig. \ref{Framework}, the proposed PDRD comprises three steps: 1) Represent each pixel by a multivariate Gaussian distribution in the latent space; 2) Select local Chebyshev neighborhood for each pixel to leverage the spatial information; 3) Obtain the detection map by manipulating the modified Wasserstein distance between the corresponding distribution of the test pixel and the average expectation of its Chebyshev neighborhood. 

\subsection{Probability Distribution Representation} 
\subsubsection{The framework} Traditional RX-based methods mainly focus on the probability distribution of background objects and assume that they obey a multivariate Gaussian distribution while the probability distribution of anomalies is hardly considered. If all training samples are modeled in a unified manner, the probability distributions can be expressed precisely. Thus the discrimination between the background and the anomalies can be evaluated directly. 

Fig. \ref{Framework} illustrates the whole framework of our proposed method. First, the training sample $x_{i} \in \mathbb R^{B}$ is fed into the encoder module $F_1$. Since our method is implemented in an unsupervised manner,  the whole pixels in HSI constitute the training samples. Next, the output of $F_1$ is fed into the probability representation part $F_2$ to generate the mean value vector and standard deviation vector. Then, we combine these two vectors with $F_3$ to produce a specific sample. Finally, the decoder module $F_4$ reconstructs the original data from the computed sample. The training is performed by minimizing the modified loss function. Specifically, the proposed network mainly comprises the four modules that take on unique responsibilities. $F_1$ denotes the encoder module that comprises three fully connected layers with 400 nodes. $F_{\text 2}$  is the probability representation part, which consists of the mean value module and the standard deviation module of the training sample. Both of the modules comprise a single layer with 20 nodes. These two outputs constitute the Gaussian distribution of the data in the latent feature space. $F_{\text 3}$ represents the sampling module, which converts the probability distribution ($\boldsymbol \mu_{i}$ and $\boldsymbol \sigma_{i}$) to a specific sample by imposing a random term $\epsilon_{i}$. $F_{\text 4}$  comprises six fully connected layers with 20 nodes. It is the decoder module to recover the original image. Both of the activation function used in the encoder and decoder network are ReLu. 	

The VAE structure seeks to construct a latent variable that reflects the features of the image in some aspects while exploring the latent representation in the probabilistic space. Moreover, the reconstructed image is generated by sampling from the latent distribution. Consequently, we regard the latent distribution as the probabilistic representation of the original data. In this paper, we creatively exploit the Gaussian distribution in the latent space to provide an explicit expression for the whole image, which significantly promotes the convenience of exploring the latent characteristic.

\subsubsection{Implementation of dimensional independence}
Different dimensions in the feature space signify different latent attributes of the original image. However, there is no explicit meaning such as size, width, or angle for each dimension corresponding to the HSI. Each pixel acts as a training sample fed to the neural network, which is substantially different from the ordinary optical image. The latent dimensions may stand for a more intrinsic structural meaning at a deeper level.

For a pixel $x_{i}$ in HSI $\mathbf{X}=\left\{x_{i}\right\}_{i=1}^{N} \in \mathbb R^{B \times N}$, we extract its distribution $q( \mathbf z)$ from $F_{\text 2}$, which consists of the mean vector $\boldsymbol \mu$ and standard deviation vector $\boldsymbol \Sigma$ of the multivariate Gaussian distribution in the latent space. However, each dimension of the distribution is not strictly independent of each other, causing impediments to compute the dissimilarity with other distributions. Hence, we strive to resolve the problem for the simplicity of implementation. For the distribution $q( \mathbf z)$ that contains $k$ latent dimensions, if all dimensions of $\mathbf z$ are mutually independent, then $q( \mathbf z)$ is equal to $\prod_{j=1}^{k} q (z_{j})$. Thus, the correlation between different dimensions can be converted to measure the difference between $q( \mathbf z)$ and $\prod_{j=1}^{k} q (z_{j})$. Our strategies that ensure mutual independence between different dimensions stem from the property.  

The loss function acts as the optimization goal of the entire network. In other words, we can reshape our loss function to optimize the framework towards the desired orientation. By adding a  weight coefficient \cite{DBLP:conf/iclr/HigginsMPBGBML17}, the expression of loss function changes to 

\begin{equation}
	\mathcal{L}=\frac{1}{N} \sum_{i=1}^{N} (E_{q_{\phi}(\mathbf{z} | \mathbf{x})}\left[\log p_{\theta}(\mathbf{x}_{i} | \mathbf{z})\right]- \beta \cdot \mathbb{KL}(q_{\phi}(\mathbf{z} | \mathbf{x}_{i}) \| p_{\theta}(\mathbf{z})))
	\label{beta loss function}
\end{equation}
Moreover, the KL divergence term in eq. \eqref{beta loss function} can be written in the form of expectation by \cite{NIPS2018_7527}

\begin{equation}
	\begin{split}
		E_{p(n)}[\mathbb{KL}(q(\mathbf z | n)|| p(\mathbf z))]=\underbrace{\mathbb{KL}(q(\mathbf{z}, n)|| q(\mathbf z) p(n))}_{\text {(i) Index-code MI }}
		\\ +\underbrace{\mathbb{KL}\left(q(\mathbf z) \| \prod_{j} q\left(z_{j}\right)\right)}_{\text {(ii) Total Correlation }}+\underbrace{\sum_{j} \mathbb{KL}\left(q\left(z_{j}\right)|| p\left(z_{j}\right)\right)}_{\text {(iii) Dimension-wise KL }}
	\end{split}
	\label{three terms}
\end{equation}
where random variable $n$ indicates the index of selected training pixel, and it obeys a uniform distribution. We can see from eq. \eqref{three terms} that the newly constructed loss function contains the term (ii) called total correlation, and the expression owns the ability to evaluate the dimensional correlation. Therefore, in order to reduce the total correlation term to the greatest extent, the optimization emphasis should be more focused on the KL divergence term rather than the expectation term in eq. \eqref{beta loss function} from a global viewpoint. The optimization balance is broken up while enhancing the value of $\beta$, and the KL divergence term receives more attention in the optimization process. Dimensional correlation declines gradually as the weight parameter $\beta$ increases, and it finally falls to a low level where it can be considered negligible. As a consequence, the dimensional independence can be satisfied theoretically.

\subsection{Selection of Chebyshev  Neighborhood}
From the process of probability representation for each pixel, we acquire the corresponding multivariate Gaussian distribution in the latent space with independent dimensions. In this way, the spectral information is fully exploited. Thus, we seek to combine the spatial information with the obtained distributions.

Given a test pixel $t$ with the two-dimensional (2D) spatial coordinate $(x,y)$, we define the point set $\mathbf V$ with $\epsilon$ Chebyshev  neighborhood by

\begin{equation}
	\mathbf V_{t}^{l}=\{(i, j)|\max (|x-i|,|y-j|)\leq \epsilon\}
	\label{set}
\end{equation}
where $l$ represents the index of points to be selected, and  $(i,j)$ denotes the coordinate of the point that satisfies the eq. \eqref{set}. 

As mentioned before, each pixel of the image corresponds to a definite multivariate Gaussian distribution in the latent feature space.  To approximate the local statistics of the test pixel, we introduce a random variable $\overline{X}_{t}\sim \mathbb {N}\left(\pmb \mu,\mathbf{\Sigma}\right)$ that reflects the average expectation of $\epsilon$ Chebyshev neighborhood for the test pixel and it can be characterized by 

\begin{gather}
	\overline{\pmb \mu}_{t} = \frac{1}{N} \sum_{i \in \mathbf{V}_{t}}{\pmb \mu_{i}} \\
	\overline{\pmb \Sigma}_{t}  = \frac{1}{N} diag\left(\sum_{i \in \mathbf{V}_{t}}{ \sigma_{1i}^{2}},\sum_{i \in \mathbf{V}_{t}}{ \sigma_{2i}^{2}},\cdots,\sum_{i \in \mathbf{V}_{t}}{ \sigma_{ki}^{2}}\right)
\end{gather}
where $\pmb \mu_{i}$ indicates the mean value vector of the distributions of $i$-th pixel in $\mathbf V_{t}$, $ {\sigma}_{ki}$ denotes the standard deviation of $k$-th dimension of $i$-th pixel in $\mathbf V_{t}$. The diagonal elements of $\mathbf{\Sigma}$ are composed by the average variance of each dimension of the probability distribution corresponding to the pixels in the $\epsilon$ Chebyshev neighborhood.

\subsection{Anomaly Detection With Modified Wasserstein Distance} 

\subsubsection{Measurement of Different Distributions}

In this step, we measure the dissimilarity of corresponding distributions between the test pixel and the average expectation of its Chebyshev neighborhood. In this paper, the concept of  Wasserstein distance is adopted to measure the distinction between two distributions, which is also employed in generative adversarial networks (GAN) to alleviate the vanishing gradient \cite{martin2017wasserstein}.

For two acquired Gaussian distributions $p = \mathcal{N}_{1}\left( \pmb {\mu}_{1},\mathbf \Sigma_{1}\right)$ and $q = \mathcal{N}_{2}\left(\pmb \mu_{2}, \mathbf \Sigma_{2}\right)$ on $\mathbb {R}^{k}$, with respective mean value vectors $\pmb \mu_{1}$ and $\pmb \mu_{2}$ $\in \mathbb{R}^{k}$,  and symmetric positive semi-definite covariance matrices $\mathbf \Sigma_{1}$ and $\mathbf \Sigma_{2} \in \mathbb{R}^{k \times k}$, the Wasserstein  distance between $p$ and $q$ is given by 
\cite{olkin1982distance}

\begin{equation}
	W\left(p, q\right)^{2}=\left\|\pmb \mu_{1}-\pmb \mu_{2}\right\|_{2}^{2}+\operatorname{\mathcal {TR}}\left(\mathbf \Sigma_{1}+\mathbf \Sigma_{2}-2\left(\mathbf \Sigma_{1} \mathbf \Sigma_{2}\right)^{\frac{1}{2}}\right)
	\label{Wasserstein distance}
\end{equation}
where $\mathcal {TR} \left(\cdot \right)$ denotes the trace of a matrix. Obviously, the computation of eq. \eqref{Wasserstein distance} is troublesome due to the complicated expression of the covariance matrix with high dimensionality.

By observing the expression of eq. \eqref{Wasserstein distance}, we know that the relationship among different dimensions of  $\mathbf z$ makes a considerable difference.
As the aforementioned results demonstrate, all dimensions of $\mathbf z$ are mutually independent, then $\mathbf \Sigma$ degenerates into diagonal matrix denoted by 

\begin{equation}
	\begin{aligned}
		\mathbf \Sigma_{1} &=  diag \left\{ \sigma_{1}^{2},\sigma_{2}^{2} , \cdots, \sigma_{k}^{2}  \right\}  \\
		\mathbf \Sigma_{2} &=  diag \left\{ \phi_{1}^{2},\phi_{2}^{2}  , \cdots , \phi_{k}^{2}  \right\}
	\end{aligned}
\end{equation}
where $\sigma_{i}$ and  $\phi _{i},(i \in \left\{1,2, \cdots k\right\})$ represent the standard deviation of each dimension for $\mathbf \Sigma_{1}$ and  $\mathbf \Sigma_{2}$, respectively. Through a series of theoretical derivations, eq. \eqref{Wasserstein distance} can be simplified to following equation:

\begin{equation}
	W\left(p, q\right)^{2}=\left\|\pmb \mu_{1}-\pmb \mu_{2}\right\|_{2}^{2}+ \left\|\pmb \sigma -\pmb \phi \right\|_{2}^{2}
	\label{Simplified distance}
\end{equation}
where $\pmb \sigma = \left[\sigma_{1}, \sigma_{2},\cdots  ,\sigma_{k} \right]^{T}$ and $\pmb \phi = \left[\phi_{1}, \phi_{2},\cdots  ,\phi_{k} \right]^{T}$. According to eq. \eqref{Simplified distance}, it is tractable to evaluate the difference between the probability distributions $p$ and $q$.

\begin{algorithm}
	\renewcommand{\algorithmicrequire}{\textbf{Input:}}
	\renewcommand{\algorithmicensure}{\textbf{Output:}}
	\caption{Anomaly detection for HSI based on PDRD}
	\label{alg:1}
	\begin{algorithmic}[1]
		\REQUIRE Training samples  $\mathbf{X} \in \mathbb R^{B \times N}$ and parameters: 1) trade-off parameter  $\beta$ and weight parameter $\gamma$; 2) Chebyshev neighborhood $\epsilon$;  3) dimensionality $k$ of latent distribution.
		\ENSURE Anomaly detection map.
		\STATE Train a  VAE neural network with modified loss function;
		\STATE Extract the probability distribution for each training sample in the latent space;
		\STATE Determine the appropriate $\epsilon$ Chebyshev neighborhood and calculate the average expectation for each pixel;
		\STATE Compute the modified Wasserstein distance between the corresponding distributions of the test pixel and the average expectation of the pixels in the $\epsilon$ Chebyshev neighborhood via eq. \eqref{Weighted Simplified distance};
		\STATE Reshape the computed distance vector to obtain the final detection map.
	\end{algorithmic}  
\end{algorithm}

\subsubsection{Modified Wasserstein distance} As eq. \eqref{Simplified distance} suggests,
the standard deviation shares equal significance with the mean value. However, in real hyperspectral scenes, the intensity of pixels from different categories oscillates much more wildly than the pixels from the same category. Hence, the difference between the anomalies and the background is biased towards the mean value rather than the standard deviation. To reshape the mutual relationship of the two terms, we attach a weight parameter $\gamma$ to the standard deviation term as follows:

\begin{equation}
	W\left(p, q\right)^{2}=\left\|\pmb \mu_{1}-\pmb \mu_{2}\right\|_{2}^{2}+ \gamma \cdot \left\|\pmb \sigma -\pmb \phi \right\|_{2}^{2}
	\label{Weighted Simplified distance}
\end{equation}
Thus, it is tractable to measure the anomalous degree of the test pixel by computing the modified Wasserstein distance between the distribution of the test pixel and the average expectation distribution of the pixels in the $\epsilon$ Chebyshev neighborhood.
The larger value of the computed Wasserstein distance indicates the greater degree of deviation from the average expectation for the test pixel. Accordingly, the pixel is more likely to become an anomaly. By reshaping the computed distance vector, we obtain the final detection map. The proposed method is elucidated in Algorithm 1.


%
%

\begin{figure}[t]
	\begin{subfigure}[t]{0.235\textwidth}
		\centering
		\includegraphics[width=1\textwidth]{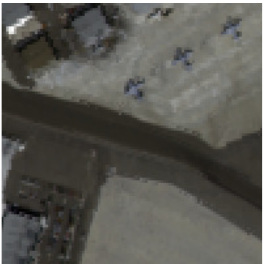}
		\subcaption{}
	\end{subfigure}
	\begin{subfigure}[t]{0.235\textwidth}
		\centering
		\includegraphics[width=1\textwidth]{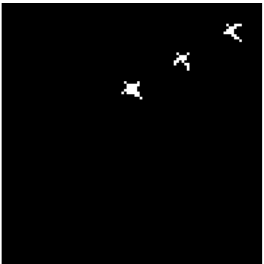}
		\subcaption{}
	\end{subfigure}
	\caption{ (a) Pseudocolor image of San Diego data set. (b) Ground Truth.  }
	\label{Sandiego}
\end{figure}

\begin{figure}[t]
	\begin{subfigure}[t]{0.235\textwidth}
		\centering
		\includegraphics[width=1\textwidth]{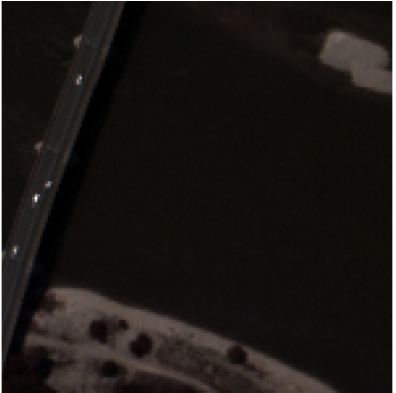}
		\subcaption{}
	\end{subfigure}
	\begin{subfigure}[t]{0.235\textwidth}
		\centering
		\includegraphics[width=1\textwidth]{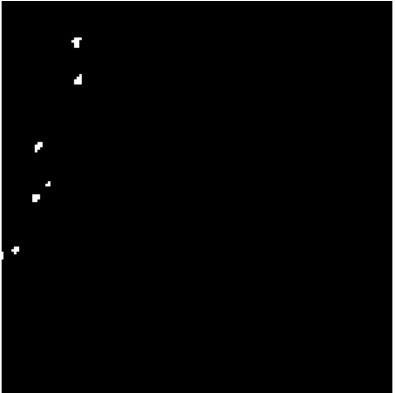}
		\subcaption{}
	\end{subfigure}
	\caption{ (a) Pseudocolor image of Pavia City data set. (b) Ground Truth.  }
	\label{Pavia}
\end{figure}

\section{Experimental results}




In this section, four real-world hyperspectral data sets are utilized to evaluate the accuracy and efficiency of the proposed method.

\subsection{Hyperspectral Data Sets}
\subsubsection{San Diego Data Set} It is a widely used data set collected by The Airborne Visible/Infrared Imaging Spectrometer (AVIRIS) sensor from San Diego, CA, USA. The spectral resolution is 10 nm, and the spatial resolution is 20 m. It has 224 bands in total ranging from 370 to 2510 nm. After eliminating the water absorption, low signal-to-noise (SNR), and bad quality bands (1-6, 33-35, 97, 107-113, 156-166, and 221-224), the remaining 189 bands are exploited for the experiments. The original image has a spatial size of 400 $\times$ 400, and a sub-image with the size of 100 $\times$ 100 in the upper left of the original image is used in our experiment. In the scene, three airplanes are considered anomalies. The main background comprises soil, parking aprons, and hangars. The pseudocolor image of San Diego data and the corresponding ground truth are shown in Fig. \ref{Sandiego}.

\subsubsection{Pavia City Data Set} The hyperspectral data set was collected by the reflective optics system imaging spectrometer (ROSIS) sensor, which covered the city center of Pavia in northern Italy. It has 205 spectral channels in total, covering the wavelength ranging from 430 to 860 nm, along with a spatial resolution of 1.3 m. The image has a geometric size of 150 $\times$ 150 pixels. The 2D visualization of the Pavia City data set and the corresponding ground truth are illustrated in Fig. \ref{Pavia}.

\begin{figure}[t]
	\begin{subfigure}[t]{0.235\textwidth}
		\centering
		\includegraphics[width=1\textwidth]{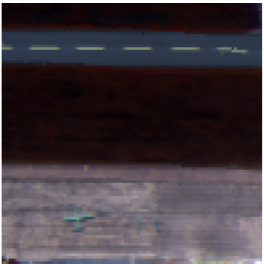}
		\subcaption{}
	\end{subfigure}
	\begin{subfigure}[t]{0.235\textwidth}
		\centering
		\includegraphics[width=1\textwidth]{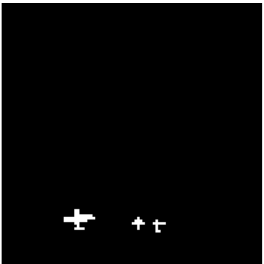}
		\subcaption{}
	\end{subfigure}
	\caption{ (a) Pseudocolor image of Gulfport data set. (b) Ground Truth.  }
	\label{Gulfport}
\end{figure}

\begin{figure}[t]
	\begin{subfigure}[t]{0.235\textwidth}
		\centering
		\includegraphics[width=1\textwidth]{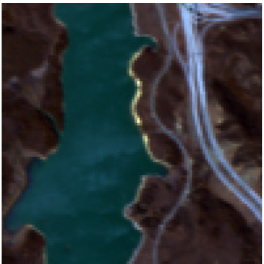}
		\subcaption{}
	\end{subfigure}
	\begin{subfigure}[t]{0.235\textwidth}
		\centering
		\includegraphics[width=1\textwidth]{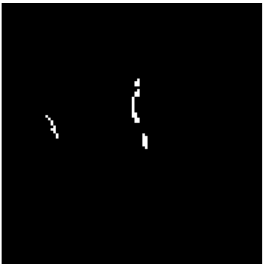}
		\subcaption{}
	\end{subfigure}
	\caption{ (a) Pseudocolor image of Jasper Ridge data set. (b) Ground Truth.  }
	\label{jasperRidge}
\end{figure}

\begin{figure*}[t]
	\begin{subfigure}[t]{0.235\textwidth}
		\centering
		\includegraphics[width=1\textwidth]{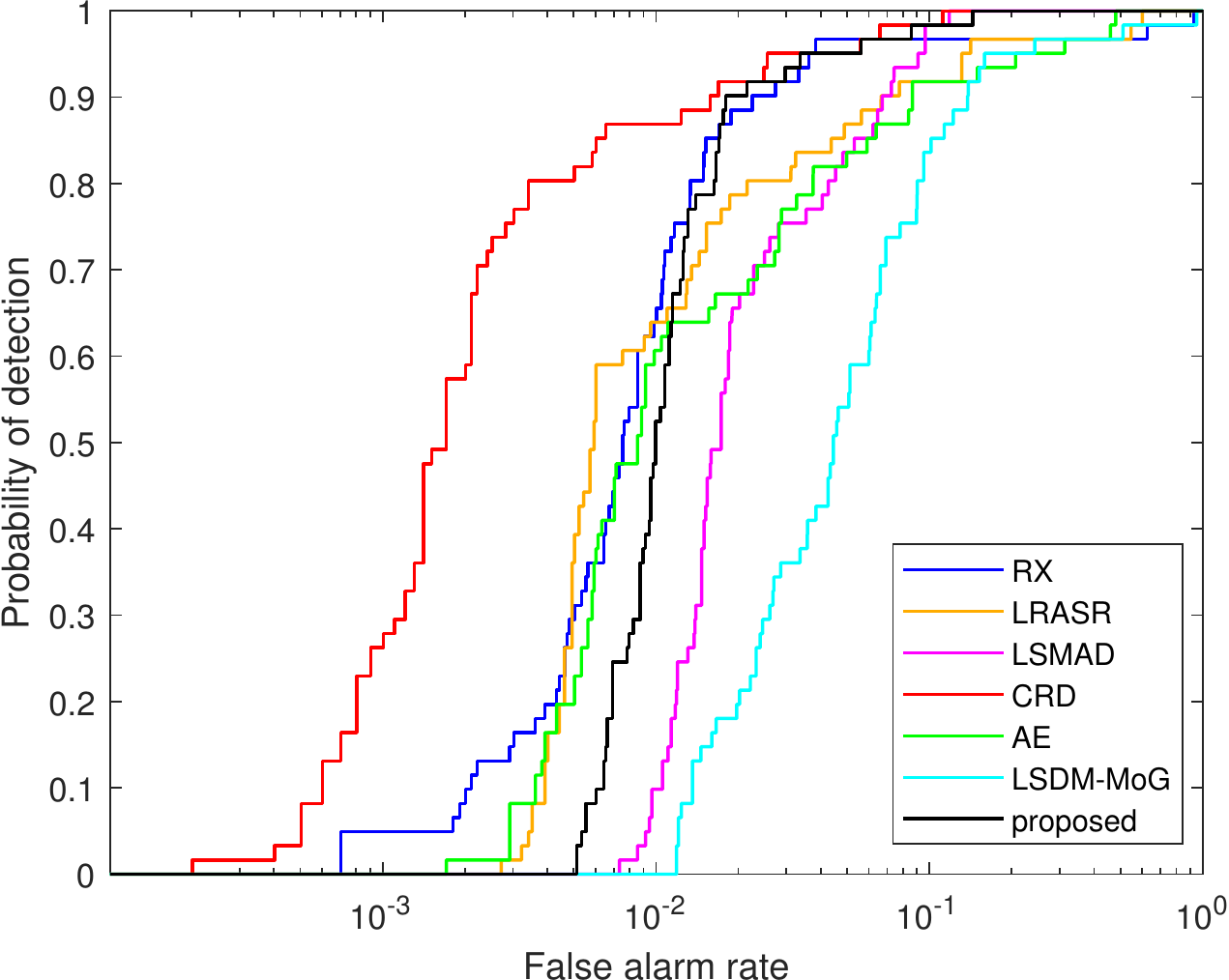}
		\subcaption{}
	\end{subfigure}
	\begin{subfigure}[t]{0.27\textwidth}
		\centering
		\includegraphics[width=1\textwidth]{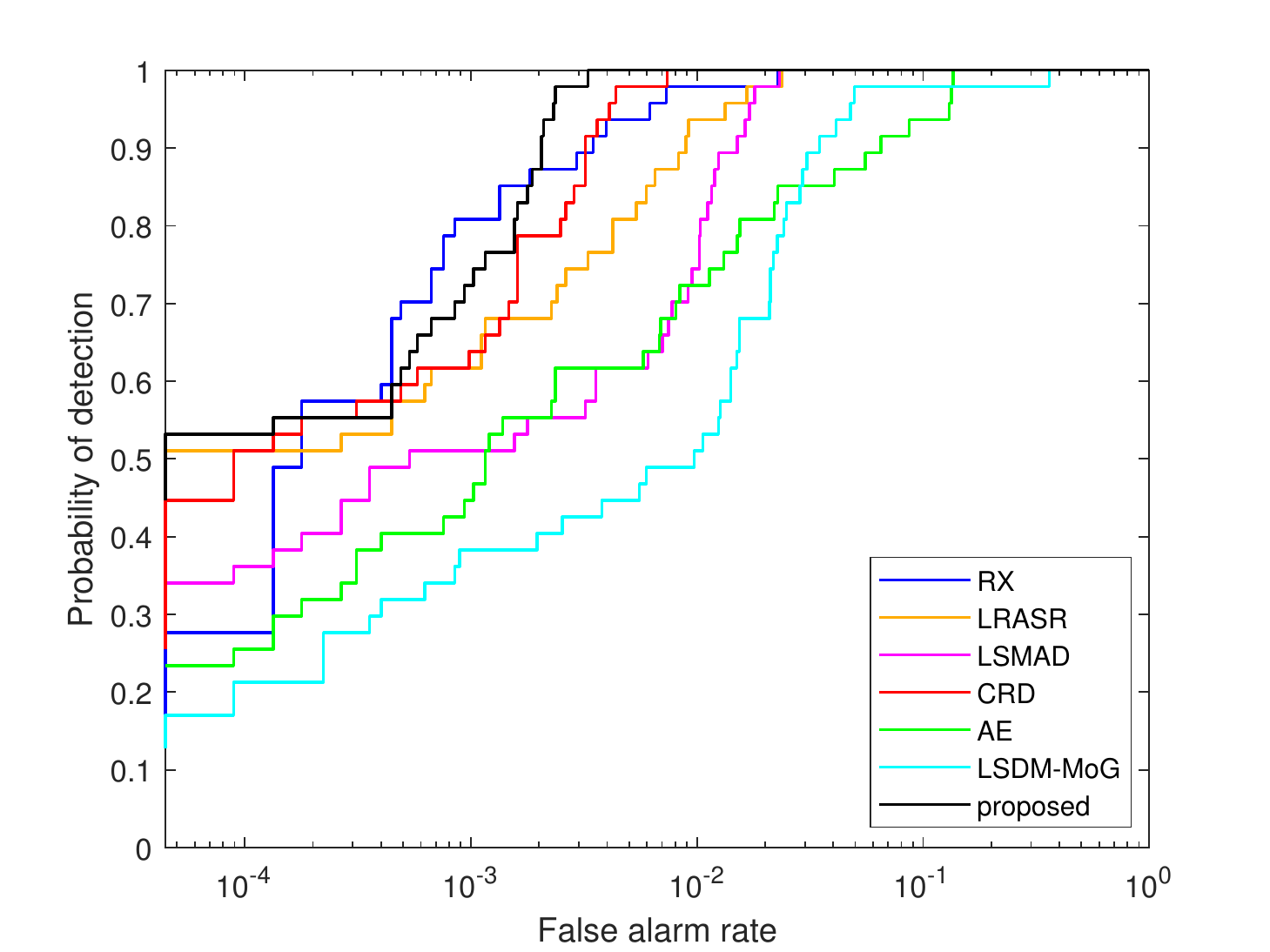}
		\subcaption{}
	\end{subfigure}
	\begin{subfigure}[t]{0.235\textwidth}
		\centering
		\includegraphics[width=1\textwidth]{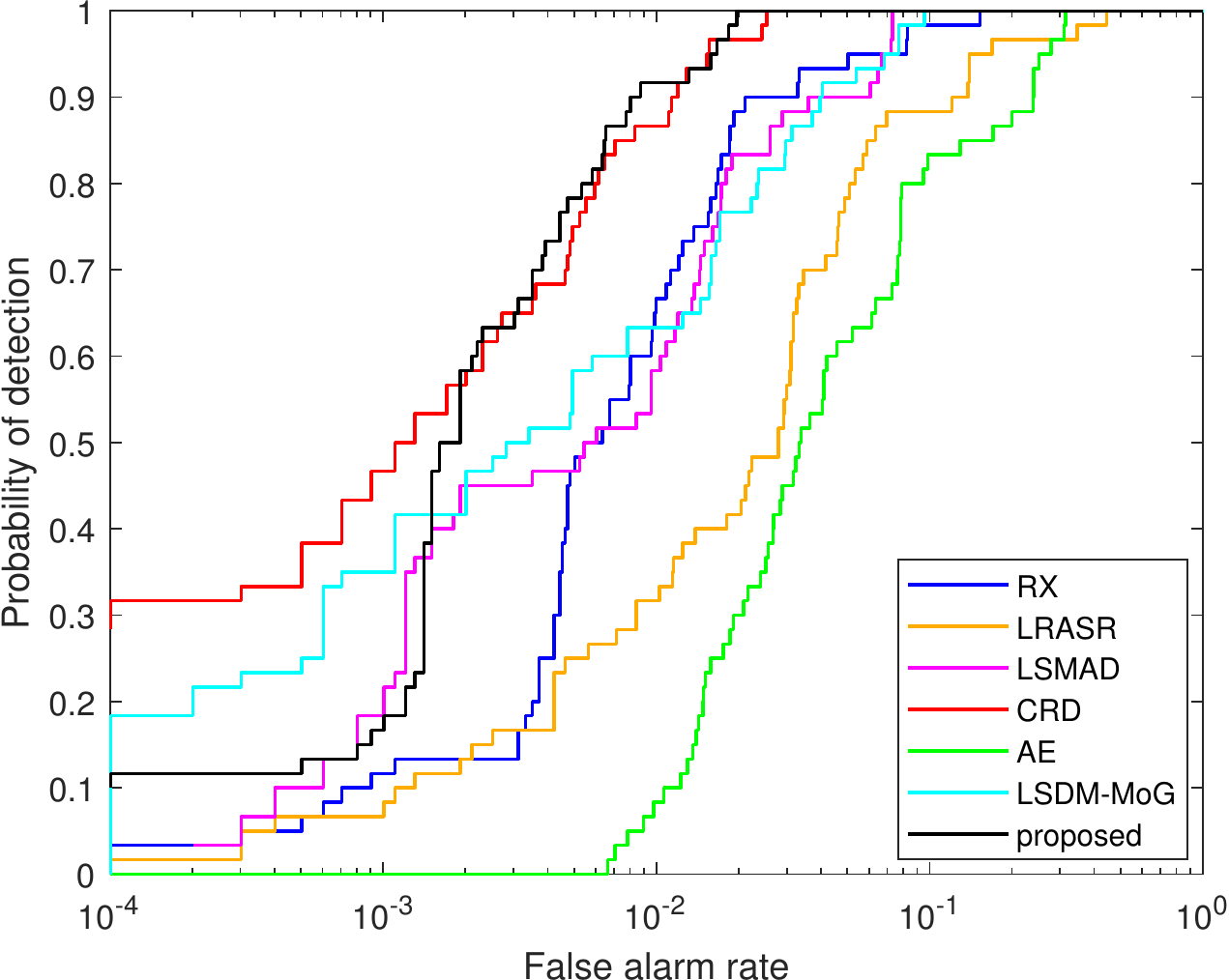}
		\subcaption{}
	\end{subfigure}
	\begin{subfigure}[t]{0.235\textwidth}
		\centering
		\includegraphics[width=1\textwidth]{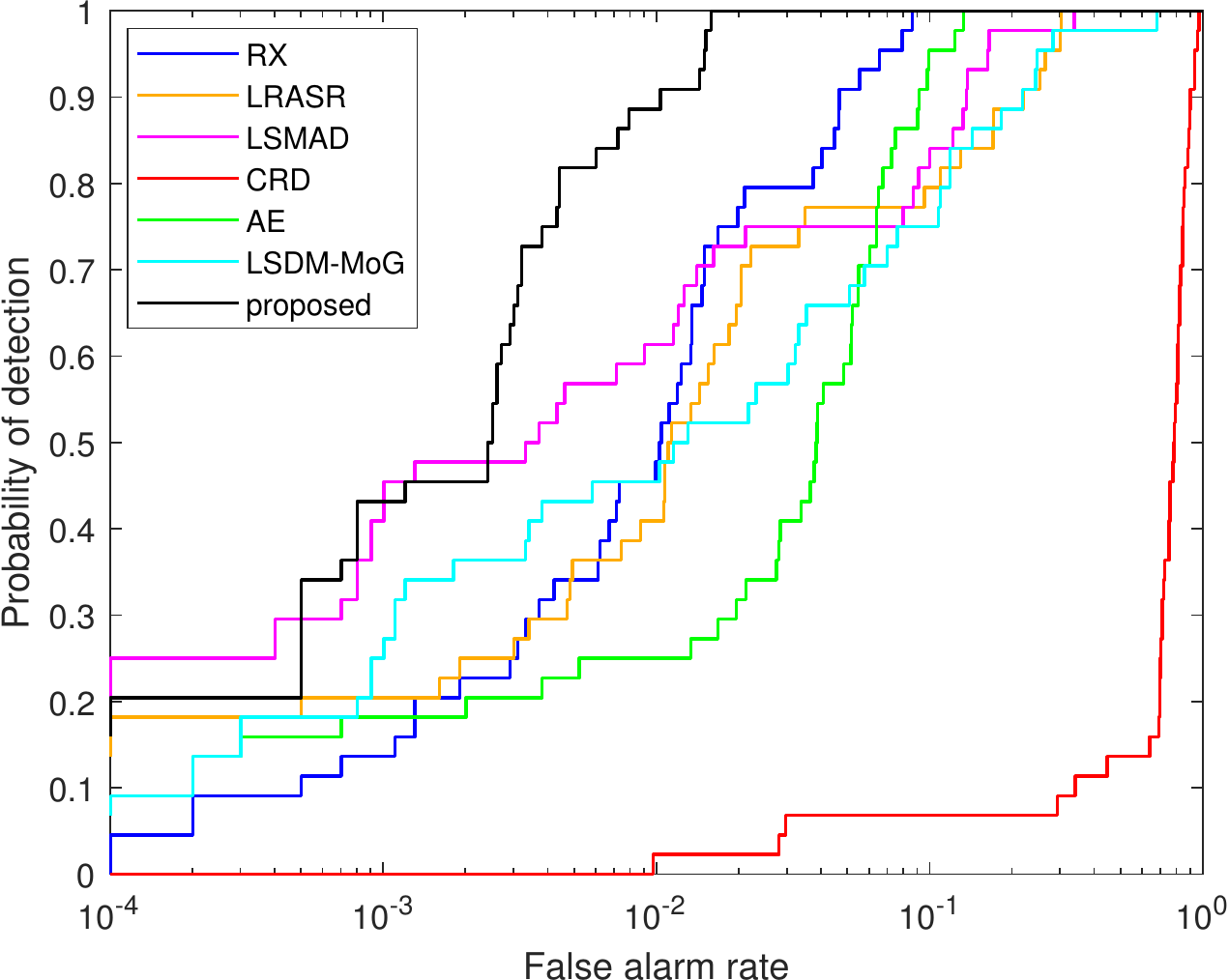}
		\subcaption{}
	\end{subfigure}
	\caption{ROC curves of $(P_{d},P_{f})$ for (a) San Diego. (b) Pavia City. (c) Gulfport. (d) Jasper Ridge. }
	\label{ ROC} 
\end{figure*}

\subsubsection{Gulfport Data Set} The data set was captured by the AVIRIS sensor over the airport area of Gulfport, MS, USA. The spatial resolution is 3.4 m, and the spatial size is 100 $\times$ 100 pixels. It contains 191 spectral channels spanning the wavelength of 550 to 1850 nm. In the scene, three planes of different sizes are regarded as anomalies. The main background of the image includes roads, vegetation, and runways. The 2D visualization of the Gulfport data set and its corresponding ground truth are depicted in Fig. \ref{Gulfport}.


\subsubsection{Jasper Ridge Data Set} It is a popular hyperspectral data set acquired by the AVIRIS sensor over Jasper Ridge in central California. The spectral resolution is up to 9.46 nm. The image records 224 channels ranging from 380 to 2500 nm. After eliminating the bands 1-3, 108-112, 154-166, and 220-224 (due to dense water vapor and atmospheric effects), the remaining 198 channels are exploited for our experiment. The original image has a spatial size of 512 $\times$ 614 pixels, and a sub-image of 100 $\times$ 100 is used for the experiment. Since the original ground truth holds the modality of probability, several contaminated objects that possess an ambiguous category are chosen as anomalies. The main background is composed of water, roads, dirt, and trees. The pseudocolor image of Jasper Ridge data and its corresponding ground truth are shown in Fig. \ref{jasperRidge}.

\begin{table}[t]
	\centering
	\caption{ AUC Scores of ($P_{d},P_{f}$) on Different Data Sets }
	\resizebox{0.5\textwidth}{!}{%
		\renewcommand{\arraystretch}{1.5}
		\setlength{\tabcolsep}{1mm}{
			\begin{tabular}{@{}cccccccc@{}}
				\toprule
				Data Set    & RX     & LRASR  & LSMAD  & CRD    & AE     & LSDM-MoG & proposed \\ \midrule
				San Diego   & 0.9634 & 0.9602 & 0.9767 & \textbf{0.9930} & 0.9566 & 0.9208   & 0.9848   \\
				Pavia City  & 0.9985 & 0.9975 & 0.9945 & 0.9989 & 0.9824 & 0.9807   & \textbf{0.9993}   \\
				Gulfport    & 0.9858 & 0.9549 & 0.9859 & 0.9962 & 0.9300 & 0.9860   & \textbf{0.9963}   \\
				Jasper Ridge & 0.9826 & 0.9467 & 0.9618 & 0.9442 & 0.9579 & 0.9368   & \textbf{0.9966}   \\
				Average     & 0.9853 & 0.9648 & 0.9797 & 0.9831 & 0.9567 & 0.9561   & \textbf{0.9942}   \\ \bottomrule
			\end{tabular}%
		}
	}
\end{table}

\subsection{Competitors And Parameter Setup}

The following state-of-the-art methods are compared with our proposed method.

1) RX detector \cite{60107} is a benchmark hyperspectral anomaly detector. It assumes that the background satisfies a multivariate Gaussian distribution. A dual-window version of RX detector is exploited in the experiments due to the high detection performance.

\begin{table}[t]
	\captionsetup{font=scriptsize}
	\centering
	\caption{ AUC Scores of ($P_{f},\tau$) on Different Data Sets }
	\resizebox{0.5\textwidth}{!}{%
		\renewcommand{\arraystretch}{1.5}
		\setlength{\tabcolsep}{1mm}{
			\begin{tabular}{@{}cccccccc@{}}
				\toprule
				Data Set    & RX     & LRASR  & LSMAD  & CRD    & AE     & LSDM-MoG & proposed \\ \midrule
				San Diego   & 0.0394 & 0.0419 & \textbf{0.0184} & 0.0205 & 0.0598 & 0.3777   & 0.0606   \\
				Pavia City  & 0.0177 & 0.0600 & \textbf{0.0087} & 0.0100 & 0.0272 & 0.0573   & 0.0213   \\
				Gulfport    & 0.0516 & 0.1057 & 0.0549 & \textbf{0.0362} & 0.0897 & 0.0954   & 0.1350   \\
				Jasper Ridge & \textbf{0.0428} & 0.1357 & 0.0585 & 0.1420 & 0.0675 & 0.1624   & 0.1043   \\
				Average     & 0.0379 & 0.0858 & \textbf{0.0351} & 0.0522 & 0.0611 & 0.1732   & 0.0803   \\ \bottomrule
			\end{tabular}%
		}
	}
\end{table}

2) LRASR \cite{7322257} adopts a background dictionary that can fully discover the implicit background structure in the latent subspace by low rank and sparse representation.  A profound separation between the background and anomalies can be achieved, and the separated anomaly part is exploited to detect anomalies.

3) LSMAD \cite{7293169} decomposes the original data into a background part, an anomaly part, and a noise part. The Mahalanobis distance that reflected background signature is computed for the following detection process.

\dr{4) CRD \cite{6876207} assumes that the background pixels can be represented by the linear combination of its spatial neighborhoods, whereas anomalies cannot.  }

5) AE \cite{10.1117/12.2195180} attempts to recover the background pixels by the structure of neural network. The anomalies hold larger reconstruction errors than the background pixels, which is the principle to distinguish the anomalies from the background.

\dr{6) LSDM-MoG \cite{9011733} models the noise component with a mixture of Gaussian distributions. The anomalies are separated from the noise components by variational Bayes. }

\dr{The parameters are set for the optimal value to make a fair comparison. For RX, the outer window size $w_{o}$ is set to 19, 25, 27, 25 for San Diego, Pavia City, Gulfport, and Jasper Ridge respectively, while the inner window size $w_{i}$ is set to 17, 19, 25, 23.} For LRASR, parameters $\beta$ and $\lambda$ are both set to 0.1. For LSMAD, it is known that the rank value $\gamma$ determines the detection accuracy. The $\gamma$ is set to 4 for all the considered data sets. For CRD, the  $w_{o}$ is set to 17, 15, 23, 25 and $w_{i}$ is set to 15, 13, 19, 7 for the corresponding data sets. For LSDM-MoG, the initial rank $l_0$ and the number of mixture Gaussian noise K are set to 60 and 4, respectively.

\begin{figure*}[ht]
	\captionsetup[subfigure]{labelformat=empty}
	\renewcommand{\thesubfigure}{\arabic{subfigure}}
	\centering
	\begin{subfigure}[t]{0.13\textwidth}
		\centering
		\includegraphics[width=1\textwidth]{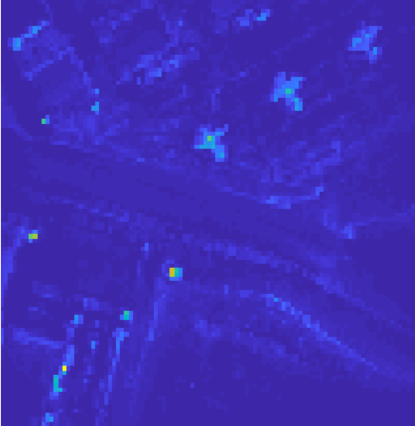}
		\subcaption{(\uppercase\expandafter{\romannumeral1}-a)}
	\end{subfigure}
	\begin{subfigure}[t]{0.13\textwidth}
		\centering
		\includegraphics[width=1\textwidth]{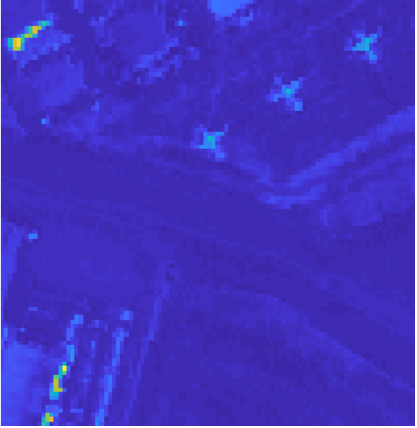}
		\subcaption{(\uppercase\expandafter{\romannumeral1}-b)}
	\end{subfigure}
	\begin{subfigure}[t]{0.13\textwidth}
		\centering
		\includegraphics[width=1\textwidth]{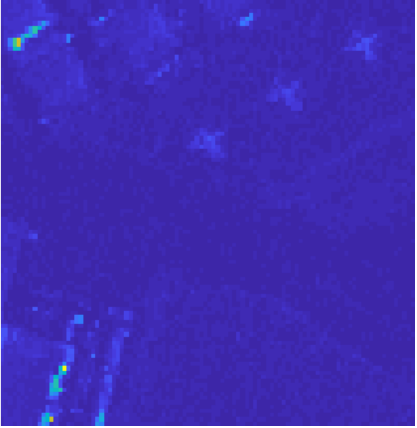}
		\subcaption{(\uppercase\expandafter{\romannumeral1}-c)}
	\end{subfigure}
	\begin{subfigure}[t]{0.13\textwidth}
		\centering
		\includegraphics[width=1\textwidth]{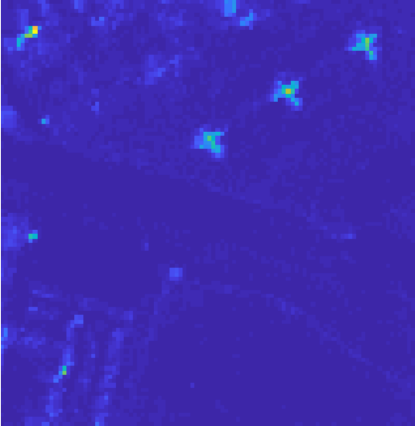}
		\subcaption{(\uppercase\expandafter{\romannumeral1}-d)}
	\end{subfigure}
	\begin{subfigure}[t]{0.13\textwidth}
		\centering
		\includegraphics[width=1\textwidth]{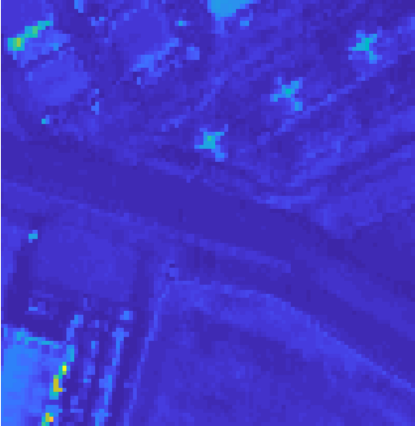}
		\subcaption{(\uppercase\expandafter{\romannumeral1}-e)}
	\end{subfigure}
	\begin{subfigure}[t]{0.13\textwidth}
		\centering
		\includegraphics[width=1\textwidth]{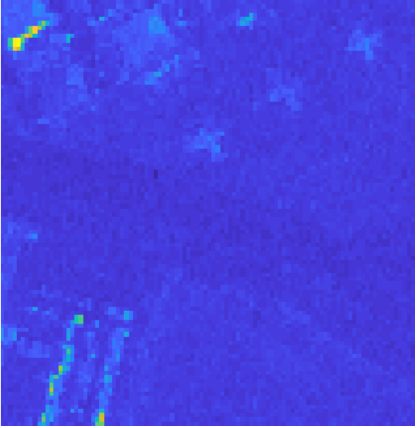}
		\subcaption{(\uppercase\expandafter{\romannumeral1}-f)}
	\end{subfigure}
	\begin{subfigure}[t]{0.13\textwidth}
		\centering
		\includegraphics[width=1\textwidth]{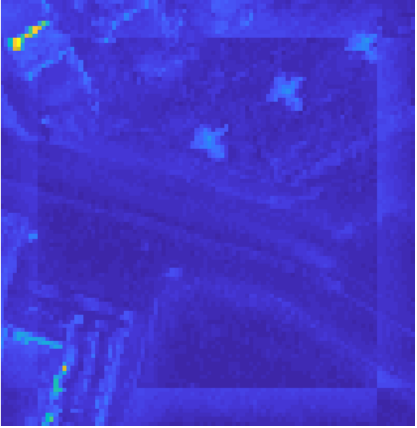}
		\subcaption{(\uppercase\expandafter{\romannumeral1}-g)}
	\end{subfigure}
	
	\begin{subfigure}[t]{0.13\textwidth}
		\centering
		\includegraphics[width=1\textwidth]{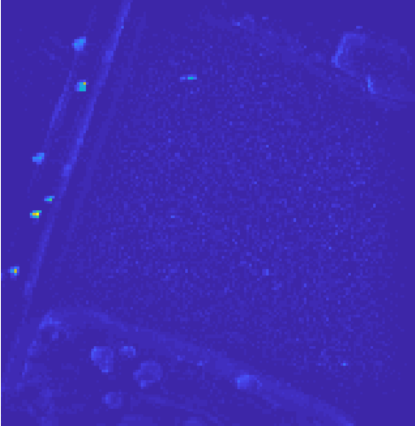}
		\subcaption{(\uppercase\expandafter{\romannumeral2}-a)}
	\end{subfigure}
	\begin{subfigure}[t]{0.13\textwidth}
		\centering
		\includegraphics[width=1\textwidth]{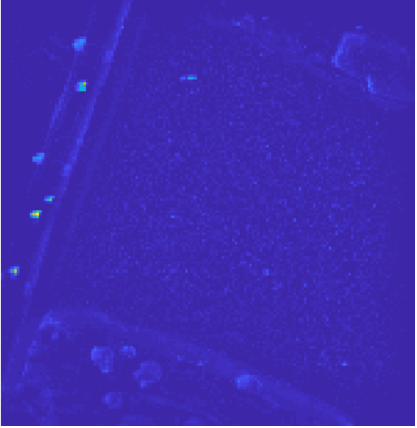}
		\subcaption{(\uppercase\expandafter{\romannumeral2}-b)}
	\end{subfigure}
	\begin{subfigure}[t]{0.13\textwidth}
		\centering
		\includegraphics[width=1\textwidth]{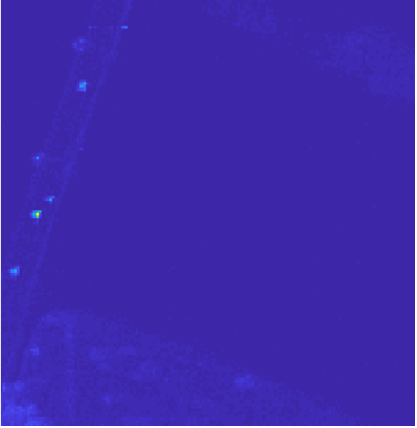}
		\subcaption{(\uppercase\expandafter{\romannumeral2}-c)}
	\end{subfigure}
	\begin{subfigure}[t]{0.13\textwidth}
		\centering
		\includegraphics[width=1\textwidth]{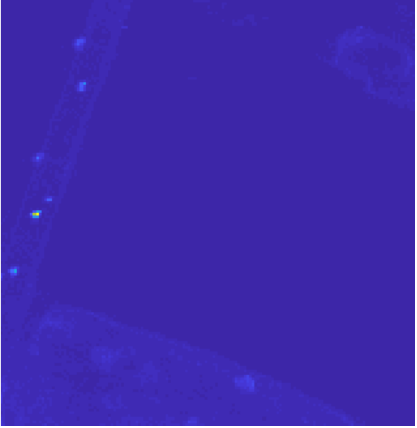}
		\subcaption{(\uppercase\expandafter{\romannumeral2}-d)}
	\end{subfigure}
	\begin{subfigure}[t]{0.13\textwidth}
		\centering
		\includegraphics[width=1\textwidth]{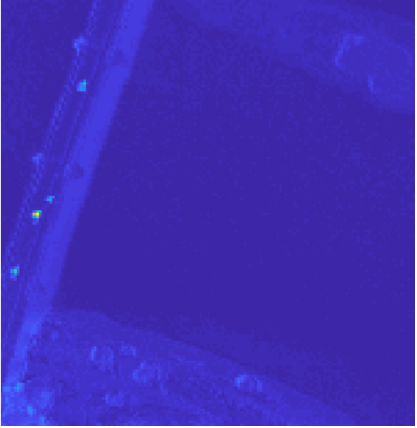}
		\subcaption{(\uppercase\expandafter{\romannumeral2}-e)}
	\end{subfigure}
	\begin{subfigure}[t]{0.13\textwidth}
		\centering
		\includegraphics[width=1\textwidth]{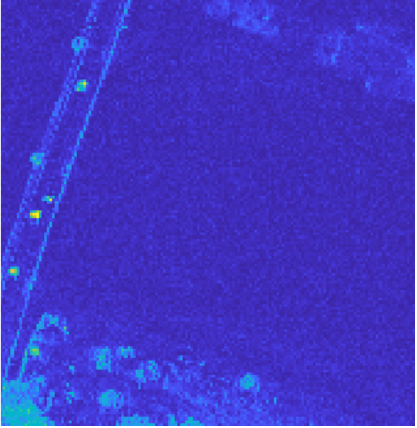}
		\subcaption{(\uppercase\expandafter{\romannumeral2}-f)}
	\end{subfigure}
	\begin{subfigure}[t]{0.13\textwidth}
		\centering
		\includegraphics[width=1\textwidth]{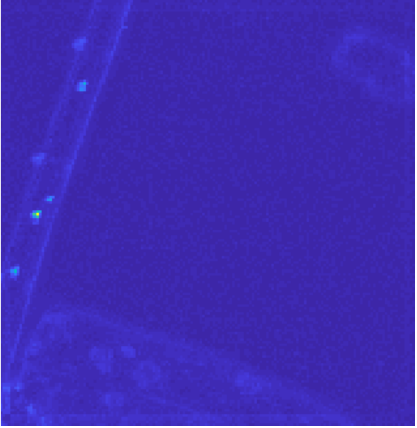}
		\subcaption{(\uppercase\expandafter{\romannumeral2}-g)}
	\end{subfigure}
	
	\begin{subfigure}[t]{0.13\textwidth}
		\centering
		\includegraphics[width=1\textwidth]{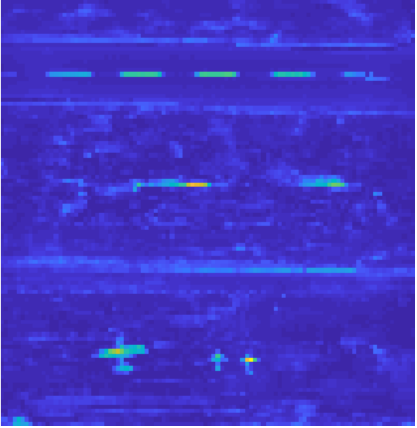}
		\subcaption{(\uppercase\expandafter{\romannumeral3}-a)}
	\end{subfigure}
	\begin{subfigure}[t]{0.13\textwidth}
		\centering
		\includegraphics[width=1\textwidth]{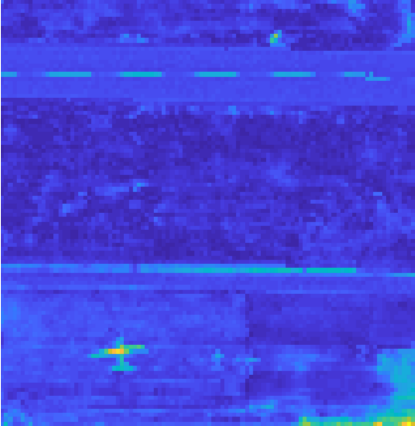}
		\subcaption{(\uppercase\expandafter{\romannumeral3}-b)}
	\end{subfigure}
	\begin{subfigure}[t]{0.13\textwidth}
		\centering
		\includegraphics[width=1\textwidth]{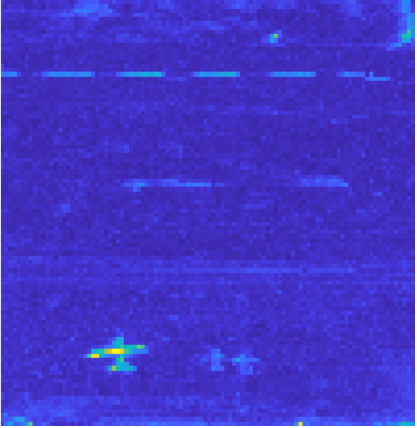}
		\subcaption{(\uppercase\expandafter{\romannumeral3}-c)}
	\end{subfigure}
	\begin{subfigure}[t]{0.13\textwidth}
		\centering
		\includegraphics[width=1\textwidth]{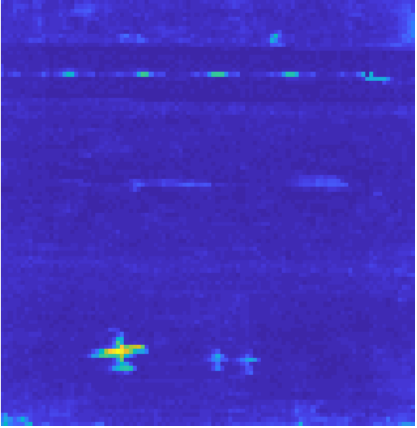}
		\subcaption{(\uppercase\expandafter{\romannumeral3}-d)}
	\end{subfigure}
	\begin{subfigure}[t]{0.13\textwidth}
		\centering
		\includegraphics[width=1\textwidth]{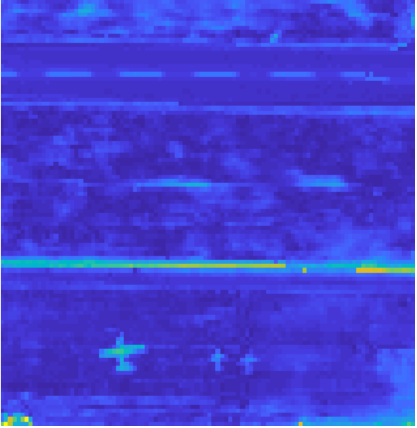}
		\subcaption{(\uppercase\expandafter{\romannumeral3}-e)}
	\end{subfigure}
	\begin{subfigure}[t]{0.13\textwidth}
		\centering
		\includegraphics[width=1\textwidth]{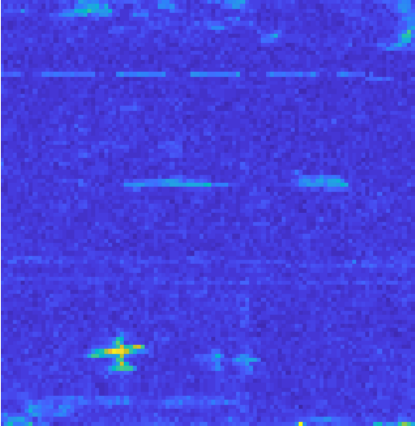}
		\subcaption{(\uppercase\expandafter{\romannumeral3}-f)}
	\end{subfigure}
	\begin{subfigure}[t]{0.13\textwidth}
		\centering
		\includegraphics[width=1\textwidth]{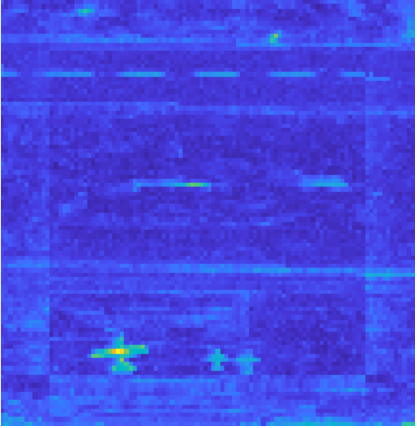}
		\subcaption{(\uppercase\expandafter{\romannumeral3}-g)}
	\end{subfigure}
	
	\begin{subfigure}[t]{0.13\textwidth}
		\centering
		\includegraphics[width=1\textwidth]{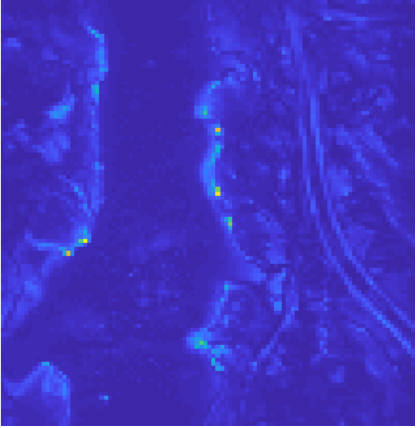}
		\subcaption{(\uppercase\expandafter{\romannumeral4}-a)}
	\end{subfigure}
	\begin{subfigure}[t]{0.13\textwidth}
		\centering
		\includegraphics[width=1\textwidth]{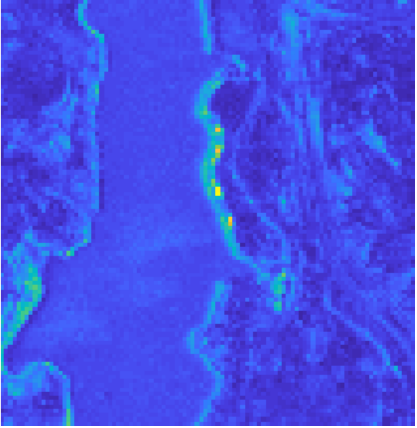}
		\subcaption{(\uppercase\expandafter{\romannumeral4}-b)}
	\end{subfigure}
	\begin{subfigure}[t]{0.13\textwidth}
		\centering
		\includegraphics[width=1\textwidth]{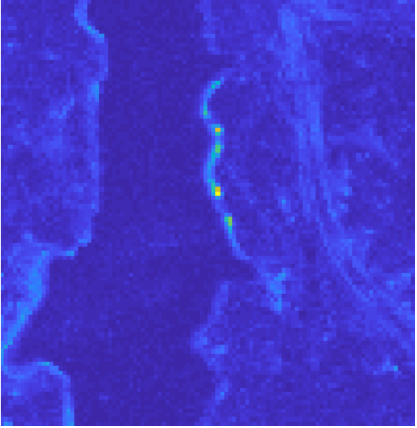}
		\subcaption{(\uppercase\expandafter{\romannumeral4}-c)}
	\end{subfigure}
	\begin{subfigure}[t]{0.13\textwidth}
		\centering
		\includegraphics[width=1\textwidth]{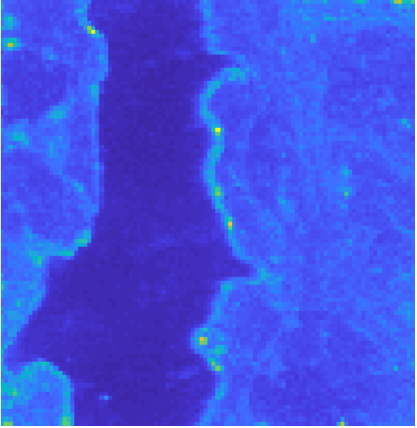}
		\subcaption{(\uppercase\expandafter{\romannumeral4}-d)}
	\end{subfigure}
	\begin{subfigure}[t]{0.13\textwidth}
		\centering
		\includegraphics[width=1\textwidth]{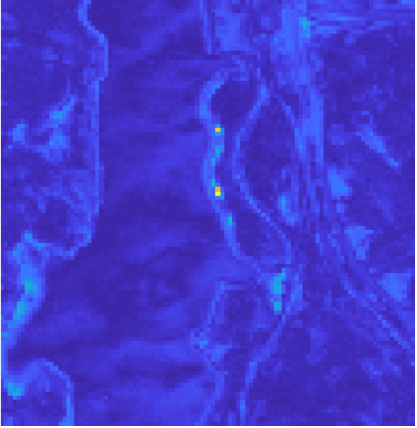}
		\subcaption{(\uppercase\expandafter{\romannumeral4}-e)}
	\end{subfigure}
	\begin{subfigure}[t]{0.13\textwidth}
		\centering
		\includegraphics[width=1\textwidth]{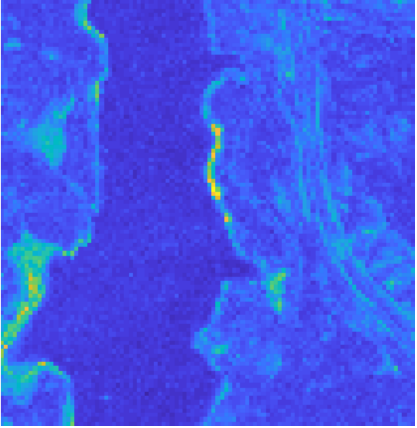}
		\subcaption{(\uppercase\expandafter{\romannumeral4}-f)}
	\end{subfigure}
	\begin{subfigure}[t]{0.13\textwidth}
		\centering
		\includegraphics[width=1\textwidth]{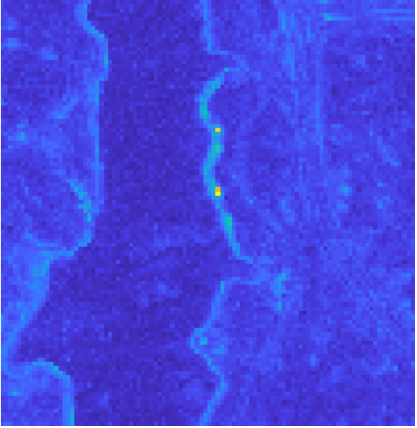}
		\subcaption{(\uppercase\expandafter{\romannumeral4}-g)}
	\end{subfigure}
	
	\caption{Visualization of the detection results.   \uppercase\expandafter{\romannumeral1} is the results of San Diego data set. \uppercase\expandafter{\romannumeral2} is the results of Pavia City data set. \uppercase\expandafter{\romannumeral3} is the results of Gulfport data set. \uppercase\expandafter{\romannumeral4} is the results of Jasper Ridge data set  . (a) RX. (b) LRASR. (c) LSMAD. (d) CRD. (e) AE.   (f) LSDM-MoG.  (g) proposed. }
	
	\label{San Diego detection} 
\end{figure*}

\subsection{Detection Performance}

Receiver operating characteristic (ROC) curve serves as a valid criterion in hyperspectral anomaly detection tasks. The superiority and effectiveness of our proposed method are illustrated by comparing the ROC curves. \dr{Fig. \ref{ ROC} illustrates the ROC curves on four data sets. }In addition, the area under the ROC curve (AUC) value of ($P_{d},P_{f}$) is used to evaluate the detection performance quantitatively. The larger AUC value manifests better detection performance. \dr{Furthermore, the AUC score of ($P_{f}, \tau$) estimates the ability of background suppression \cite{9082180}. The lower value indicates higher performance of background suppression. Tables \uppercase\expandafter{\romannumeral1} and \uppercase\expandafter{\romannumeral2} show the AUC scores of ($P_{d},P_{f}$) and ($P_{f},\tau$) on the four real HSIs.}

\begin{figure*}[t]
	\begin{subfigure}[t]{0.33\textwidth}
		\centering
		\includegraphics[width=1\textwidth]{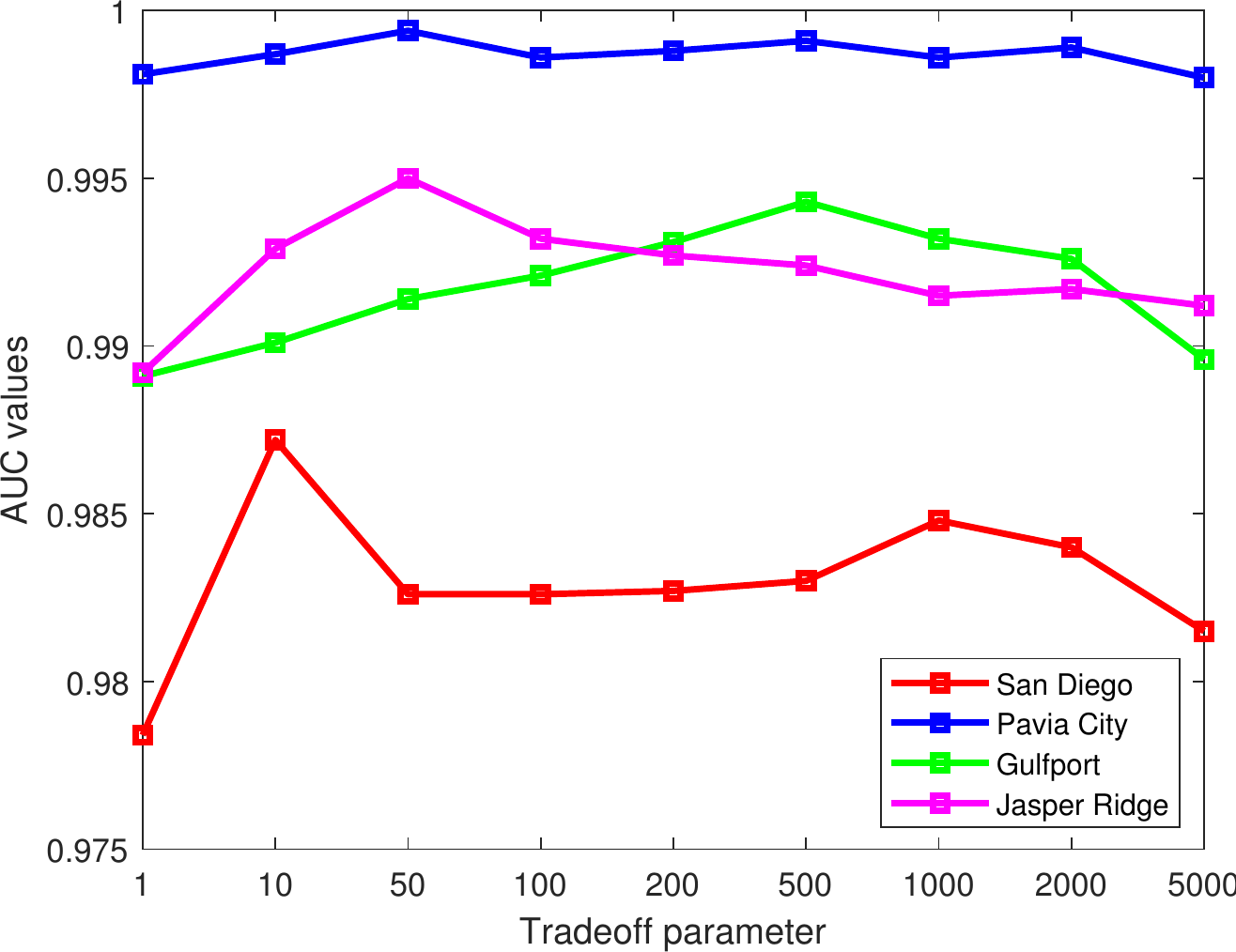}
		\subcaption{}
	\end{subfigure}
	\begin{subfigure}[t]{0.33\textwidth}
		\centering
		\includegraphics[width=1\textwidth]{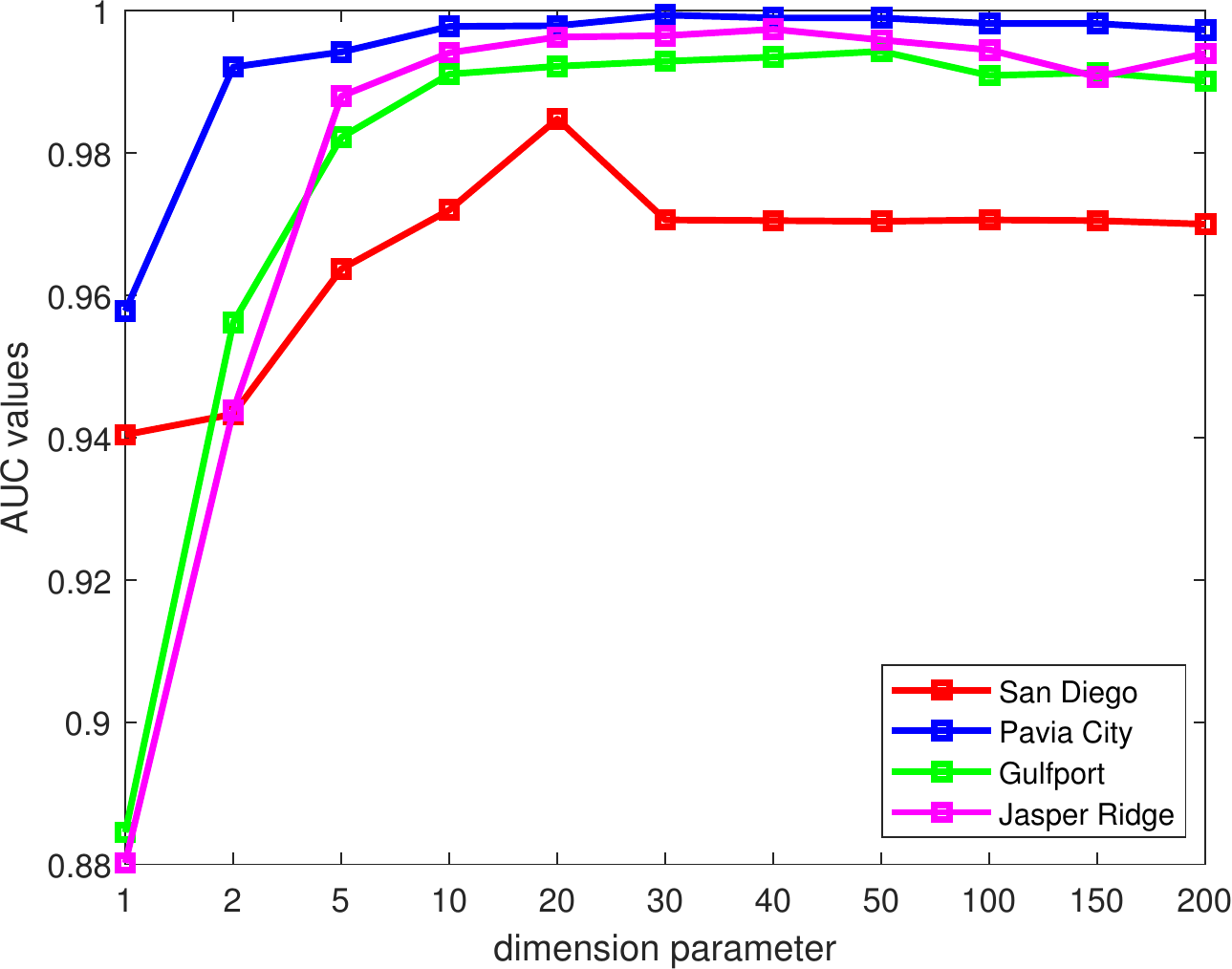}
		\subcaption{}
	\end{subfigure}
	\begin{subfigure}[t]{0.33\textwidth}
		\centering
		\includegraphics[width=1\textwidth]{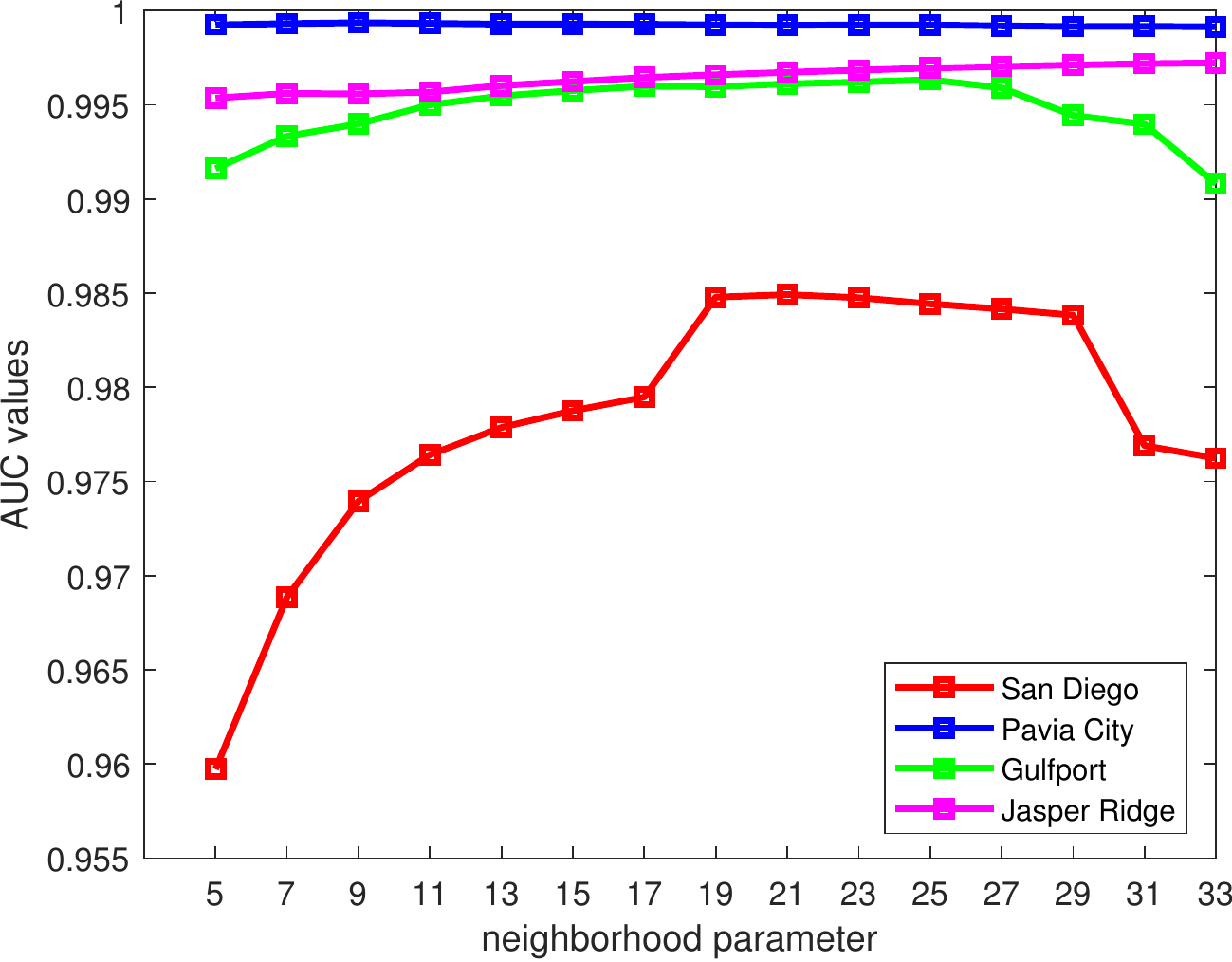}
		\subcaption{}
	\end{subfigure}
	\caption{AUC values of ($P_{d},P_{f}$) with the change of different parameters. (a) tradeoff parameter $\beta$. (b) dimension parameter $k$. (c) neighborhood parameter $\epsilon$. }
	\label{ Parameter} 
\end{figure*}

\subsubsection{Results of San Diego data} Fig. \ref{San Diego detection}-\uppercase\expandafter{\romannumeral1} displays the visualization of detection results on the San Diego data set. Concerning the San Diego data set, the general background is relatively more complex than others. The LSDM-MoG method holds the worst performance among these methods. Neither it detects many anomalies nor excludes enough background pixels. The RX and LRASR methods fail to detect a certain amount of anomalies, causing poor detection performance. The LSMAD method integrates matrix decomposition theory with the classical RX detector and employs the noise term in the matrix decomposition process. However, both the anomalies and the background are suppressed. Consequently, the anomalies are not evident enough to be distinguished from the background pixels. Moreover, the AE method detects most of the anomalies, while several background objects whose spectral signature is similar to the airplane anomalies are also detected as anomalies. As a result, the algorithm receives bad detection performance. In contrast, our proposed method achieves outstanding performance, and the detected anomalies are distinct.

ROC curves are shown in Fig. \ref{ ROC}(a) to demonstrate the effectiveness of our proposed method. The ROC curve of our proposed method has a quick responding speed when the false alarm rate is nearly $\text10^{\text{-2} }$. When the detection of probability reaches 1, the false alarm rate is better than some other methods. \dt{Notably, our proposed method is the fastest to reach the position where the probability of detection is equal to 1. It means the minimal regularized intensity of the detected anomalies in our proposed method is more significant than that of other detectors, indicating the deeper separation between the anomalies and the background pixels.}

The AUC scores of ($P_{d},P_{f}$) on the San Diego data set are shown in Table \uppercase\expandafter{\romannumeral1}. Evidently, our proposed method achieves the second-largest AUC value of ($P_{d}, P_{f}$) and yields better detection performance than other methods by a large margin except for CRD. \dt{We can conclude from the AUC values LRASR and LSMAD  that it is difficult to reconstruct the anomaly part under the complex background condition. The local method like LRX gets a relative high AUC value, but high false alarm rate prevents it from growing further. AE method fails to explore the intrinsic structure of background distribution, causing a low AUC value.}

\subsubsection{Results of Pavia City data} The visualization of detection results on the Pavia City is shown in Fig. \ref{San Diego detection}-\uppercase\expandafter{\romannumeral2}. \dt{We can observe that both RX and RPCA miss a few scattered anomalies as the total intensity of their detection maps is very low.} RX and LSDM-MoG highlights both the background and the anomalies, leading to a high false alarm rate. The detected anomalies of LRASR contains background components more or less, due to the large detection value of the background pixels. LSMAD faces the dilemma that the intensity of some anomalies is very low. CRD misses several targets, generating lower detection performance. Furthermore, the detection performance of AE attenuates because several bridge objects are detected as anomalies. Our proposed method fully utilizes global information of the data and incorporates the strength of local neighborhood into the detection process to enhance the saliency of anomalies.

\begin{table}[t]
	\captionsetup{font=scriptsize}
	\centering
	\caption{ Execution Time (in Seconds) on Different Data Sets }
	\resizebox{0.5\textwidth}{!}{%
		\renewcommand{\arraystretch}{1.5}
		\setlength{\tabcolsep}{1mm}{
			\begin{tabular}{@{}cccccccc@{}}
				\toprule
				Data Set    & RX     & LRASR  & LSMAD  & CRD    & AE     & LSDM-MoG & proposed \\ \midrule
				San Diego   & 58.00 & 24.33  & 8.62  & 9.67   & 51.28 & 4.91     & \textbf{2.63}   \\
				Pavia City  & 84.06 & 133.00 & 10.83 & 686.00 & 56.11 & 11.04    & \textbf{3.12}   \\
				Gulfport    & 86.30 & 23.04  & 8.40  & 15.88  & 36.56 & 6.72     & \textbf{6.19}   \\
				Jasper Ridge & 86.02 & 27.24  & 11.54 & 583.77 & 30.93 & 6.25     & \textbf{3.60}   \\
				Average    & 78.60 & 51.89  & 9.85  & 323.80 & 43.72 & 7.23     & \textbf{3.89}  \\ \bottomrule
			\end{tabular}%
		}
	}
\end{table}

ROC curves are plotted for the Pavia City to compare the detection performance qualitatively, as Fig. \ref{ ROC}(b) shows. \dt{ Obviously, the ROC curve of our proposed method locates in the upper-left corner of the figure, which embodies prominent detection performance.} \dr{Notably, our proposed method is the fastest to reach the position where the probability of detection is equal to 1. It means the minimal intensity of the detected anomalies in our proposed method is more significant than that of other detectors.} In the preliminary stage, the probability of detection is over 0.5, which dominates the value among these methods. When the probability of detection approaches 1, the false alarm rate is less than $\text10^{\text{-2} }$, which precedes other methods by a large margin.

\begin{table}[t]
	\captionsetup{font=scriptsize}
	\centering
	\caption{ AUC Scores for Component Analysis }
	\resizebox{0.5\textwidth}{!}{%
		\renewcommand{\arraystretch}{1.5}
		\setlength{\tabcolsep}{1mm}{
			\begin{tabular}{@{}cccccccc@{}}
				\toprule
				Component    & San Diego     & Pavia City  & Gulfport  & Jasper Ridge    & Average  \\ \midrule
				PDRD without PR   & 0.6753 & 0.9350  & 0.6432 & 0.6812   & 0.7337  \\
				PDRD without CN  & 0.9783 & 0.9753 & 0.9868 & 0.9764 & 0.9792  \\
				PDRD without MLF    & 0.9751 & 0.9971  & 0.9906  & 0.9925  & 0.9888   \\
				PDRD & \textbf{0.9844} & \textbf{0.9991}  & \textbf{0.9958} & \textbf{0.9968} & \textbf{0.9940}\\ \bottomrule
			\end{tabular}%
		}
	}
\end{table}

The background condition of Pavia City data set is not as complicated as that of the San Diego data. Hence, all methods yield excellent detection performance. The AUC value of ($P_{d},P_{f}$) in our proposed method is 0.9993, which is very near to 1. The value is convincing because it is consistent with the aforementioned analysis outcomes concerning the visualization of detection results and ROC curves. 

\subsubsection{Results of Gulfport data} Fig. \ref{San Diego detection}-\uppercase\expandafter{\romannumeral3} shows the visualization of detection maps on the Gulfport image. \dt{Both GRX and RPCA fail to detect a set of anomalies due to the diversity of background types.}RX, CRD, and LSDM-MoG incorporate a number of the runway objects into anomalies, as we can observe that several horizontal stripes are significant. LRASR erroneously detects large areas of pixels located in the lower-right of the original image as anomalies, causing bad detection performance. Besides, some background pixels of LSMAD are salient than anomalies, leading to a low detection rate. AE regards road objects as anomalies, and we can see a horizontal line in the detection result.

ROC curves are plotted as shown in Fig. \ref{ ROC}(c) for the Gulfport data to confirm the detection performance qualitatively. The ROC curve of our proposed method keeps at a high level since the false alarm rate is over  $\text10^{\text{-3} }$, and the tendency of growth holds quite steep. When the probability of detection of our proposed method reaches 1, the corresponding value of other methods is far from our proposed method. 

For the Gulfport data set, our method makes a  considerable improvement towards AUC scores of ($P_{d},P_{f}$) compared with other methods, which can be owed to the utilization of both global and local information of the data. The results are consistent with the evaluation of both ROC curves and the visualization of detection maps. As a consequence, we conclude that our algorithm achieves  better performance.

\subsubsection{Results of Jasper Ridge data} Fig. \ref{San Diego detection}-\uppercase\expandafter{\romannumeral4} displays the detection results on Jasper Ridge data. LSDM-MoG gets the worst performance among these methods. It detects most of the anomalies, but several background pixels are highlighted. CRD, LSMAD, and AE yield better results than LSDM-MoG. However, they all miss detecting plenty of anomalies. LRASR has a high false alarm rate at the lower-left part of the image. Since the category of each object is not deterministic for this data set, the effect of constructing a background dictionary decreases. Therefore, the detection performance of LRASR diminishes rapidly compared to other data sets. Our method and RX have achieved better performance than other methods. Moreover, the results demonstrate slight superiority for our proposed method because the regularized intensity of anomalies is higher than RX.

ROC curves are shown in Fig. \ref{ ROC}(d)  to evaluate the detection results from the qualitative perspective. As we can see, the ROC curve of our method lies almost in the upper-left corner of the image. When the false alarm rate is over $\text10^{\text{-2} }$, the probability of detection is near to 1, which leads the performance among these methods to a great extent.

In line with the conclusion regarding the visualization of detection results and ROC curves, our method holds the best detection performance regarding AUC scores of ($P_{d},P_{f}$). The effectiveness and superiority stem from the consideration of the intrinsic probability distribution of every single pixel. As a consequence, the implicit structure of both anomalies and the background is well-recovered. \dt{The minor difference between the anomalies and the background is enhanced, which is favorable for detecting anomalies.}

\subsection{Parameter Analysis}

\subsubsection{Weight Parameter $\gamma$ and Tradeoff Parameter $\beta $ } The weight parameter $\gamma$ determines the biased degree between the mean value and the standard deviation. After multiple experiments have been conducted, all of them reach the same conclusion that only if $\gamma $ is set to 0 can our model reach powerful detection performance. Under the circumstance, the standard deviation acts as interference in our detection task, which suggests only the mean value of the distribution influences the discrimination between the background and the anomalies. Hence,  $\gamma$ is set to 0 for all following experiments.

The tradeoff parameter $\beta$ is used to balance the weight of reconstruction error and KL divergence term in the VAE loss function. The setting of $\beta$ is significant in our experiment because the mutual independence of different dimensions depends on $\beta$. The experimental results further verify our aforementioned theoretical analysis. As we can observe from Fig. \ref{ Parameter}, the AUC value grows as $\beta$ increases in the preliminary stage. After the AUC value reaches the maximum, it declines slowly from a general perspective. Specifically, with the increase of $\beta$, although the distribution in the latent space updates towards the direction that each dimension is mutually independent, the emphasis of the loss function in eq. \eqref{beta loss function} gradually tilts toward the KL divergence term, leading to the decline of the model reconstruction ability. The optimal parameter is acquired by trial and error, and the critical point is to unearth a balance between the reconstruction error and the KL divergence term.

The dimensionality $k$ in the latent space is set to 20, while the local neighborhood $\epsilon$ is set to 19. Since the value of the KL divergence term is small, the range of the tradeoff parameter $\beta$  we set is $\left\{1, 10, 50, 100, 200, 500, 1000, 2000, 5000\right\}$ in the experiment. When  $\beta$ is equal to 1, the neural network represents the vanilla VAE. Moreover, the influence of $\beta$ on four different hyperspectral data sets is not identical. The optimal setting values of $\beta$ are 10, 50, 500, and 50 for San Diego, Pavia City, Gulfport, and Jasper Ridge data, respectively.

\subsubsection{Dimensionality $k$ of Latent Variable} The distribution of latent variable $z$ signifies the intrinsic structure of original data in the feature space. The setting of $k$ also deserves considerable attention.  We can observe from Fig. \ref{ Parameter}(b) that the AUC value proliferates as $k$ increases at the very beginning. After the AUC value reaches the maximum, it keeps at a high level or drops slightly. The phenomenon can be explained from the perspective of dimensional correlations. When $k$ is too small, massive redundancy and high correlations exist between different dimensions. As a result, sufficient information cannot be fully extracted by low dimensional probability representation in the latent space. On the contrary, provided that $k$ is large enough or larger than the band number of HSI, massive redundant dimensions emerge.

Both $\beta$ and $\epsilon$  remain unchanged when we optimize $k$. Considering the actual band number of HSI, the range of parameter $k$ is set to $\left\{1, 2, 5, 10, 20, 30, 40, 50, 100, 150, 200\right\}$. The experimental results indicate that the optimal setting values of $k$ are 20, 30, 50, and 40 for San Diego, Pavia City, Gulfport, and Jasper Ridge data, respectively. 

\subsubsection{Chebyshev neighborhood $\epsilon$} The Chebyshev neighborhood $\epsilon$ determines the range of local information we utilize from the original hyperspectral image. Fig. \ref{ Parameter}(c) demonstrates that when $\epsilon$ increases, the AUC value enhances at the very beginning. The AUC value diminishes as  $\epsilon$ further increases after it reaches the maximum. When $\epsilon$ is small, the neighborhood only covers a small portion of adjacent areas, thus the local information cannot be exploited thoroughly.  When $\epsilon$ is too large, the adjacent areas contain multiple unexpected background categories. Moreover, the execution time increases quickly as $\epsilon$ grows.

The $\beta$ and $k$ maintain fixed when exploring the optimal setting of $k$. The detection results displayed in Fig. \ref{ Parameter}(c) reveal that the best of $\epsilon$ values we set are 21, 9, 25, and 31 for San Diego, Pavia City, Gulfport, and Jasper Ridge data, respectively. 

\subsubsection{Parameters of Neural Network} The learning rate $\alpha$ is crucial to the training procedure of a neural network.   The appropriate setting value of $\alpha$ can accelerate the training process and promote the performance of the neural network. In our experiment, $\alpha$ is set to $\text 10^{\text{-5} }$ for Jasper Ridge data set and $\text 10^{\text{-3} }$ for other three data sets.

The batch size of a neural network influences the learning accuracy and updating speed, and the optimal value is derived from trial and error with prior knowledge. It is set to 16, 32, 16, and 32  for San Diego, Pavia City, Gulfport, and Jasper Ridge data sets, respectively. 

\subsection{Execution Time}

In this part, we compare the execution time of the aforementioned detectors. The experiments are conducted on a computer with a 64-bit Intel i7-8700 CPU of 3.2 GHz on Windows 10. \dr{To make a fair comparison, all methods are implemented with CPU.}  The average execution time of the state-of-the-art methods on different data sets can be observed from Table \uppercase\expandafter{\romannumeral3}. As we can see, our proposed method takes the least time among these methods. RX and CRD are much slower as the sliding windows consume considerable time.  As for matrix decomposition-based methods, LRASR costs much more time than LSMAD due to the optimization of the background dictionary. AE also consumes much time because the reconstruction error cannot be minimized to the expected level until numerous iterations.

Notably, our proposed method not only achieves high detection performance, but also obtains high computational efficiency simultaneously compared to other methods. Low computational cost and high detection accuracy make it possible to implement the method into practical applications in the technological industry.

\subsection{Ablation Study}
To evaluate the effect of each part on the final detection result, including probabilistic representation (PR), Chebyshev neighborhood (CN), and modified loss function (MLF), we conduct an ablation study on four real hyperspectral data sets by using different combinations of the network components. In the first scenario, we eliminate the PR part, thus the architecture degenerate into the VAE network. Evidently, the AUC score decreases significantly on four data sets without the PR module, which indicates the validity of PR module. With the removal of CN in the second scenario, the average AUC score is reduced by 1.5$\%$, which manifests the usefulness of CN. When MLF is removed in the third scenario, the  AUC score declines by 0.5$\%$,  which embodies the effectiveness of MLF. From a comprehensive point of view, it can be observed that the average  AUC score of the models without PR, CN, and MLF are smaller than PDRD, which signifies the combination of Gaussian probabilistic representation, local Chevbyshev neighborhood and the modification of loss function is effective for anomaly detection.

\section{Conclusion}
In this paper, we propose a novel PDRD for hyperspectral anomaly detection. We exploit multivariate Gaussian distributions to represent the whole image in the latent space from a probabilistic perspective.  To take advantage of the spatial information, we introduce the $\epsilon$ Chebyshev neighborhood and average expectation to estimate the local statistics. Finally, the modified Wasserstein distance serves as the evaluation criterion for anomaly detection. The distinction between the corresponding distributions of the test pixel and the average expectation in the neighborhood is computed to acquire the anomalous degree for each pixel. Experiments on four real hyperspectral scenes demonstrate the accuracy and efficiency of our proposed method compared to the state-of-the-art methods. In future work, we intend to combine the probability representation strategy with other neural networks.

\section*{Acknowledgement}
This work was supported in part by the National Nature Science Foundation of China under
Grant 61671408, and in part by the Joint Fund of the Ministry of Education of China under Grant 6141A02022362.

\printcredits

\bibliographystyle{model1-num-names}
\bibliography{cas-refs}


%
%

\end{document}